%% file: egpaper_final.tex
\documentclass[10pt,twocolumn,letterpaper]{article}
\pdfoutput=1
\usepackage{cvpr}
\usepackage{times}
\usepackage{epsfig}
\usepackage{graphicx}
\usepackage{amsmath}
\usepackage{amssymb}
\usepackage{subfig}
\usepackage{multirow}
\usepackage{algorithm}
\usepackage[noend]{algpseudocode}
\usepackage [english]{babel}
\usepackage [autostyle, english = american]{csquotes}
\MakeOuterQuote{"}
\usepackage{dblfloatfix}
\usepackage{booktabs}
\newcommand{\ra}[1]{\renewcommand{\arraystretch}{#1}}

\usepackage[breaklinks=true,bookmarks=false]{hyperref}
\hypersetup{
    colorlinks=true,
    linkcolor=blue,
    filecolor=blue,      
    urlcolor=blue,
}

\cvprfinalcopy 

\def\cvprPaperID{6308} 

\ifcvprfinal\fi
\begin{document}

\title{"Looking at the right stuff" - Guided semantic-gaze for autonomous driving}

\author{Anwesan Pal\\
UC San Diego\\
{\tt\small a2pal@eng.ucsd.edu}
\and
Sayan Mondal\\
UC San Diego\\
{\tt\small samondal@eng.ucsd.edu}
\and
Henrik I. Christensen\\
UC San Diego\\
{\tt\small hichristensen@eng.ucsd.edu}
}

\maketitle
\thispagestyle{empty}

\begin{abstract}
In recent years, predicting driver's focus of attention has been a very active area of research in the autonomous driving community. Unfortunately, existing state-of-the-art techniques achieve this by relying only on human gaze information, thereby ignoring scene semantics. We propose a novel Semantics Augmented GazE (\textbf{SAGE}) detection approach that captures driving specific contextual information, in addition to the raw gaze. Such a combined attention mechanism serves as a powerful tool to focus on the relevant regions in an image frame in order to make driving both safe and efficient. Using this, we design a complete saliency prediction framework - \textbf{SAGE-Net}, which modifies the initial prediction from SAGE by taking into account vital aspects such as distance to objects (depth), ego vehicle speed, and pedestrian crossing intent. Exhaustive experiments conducted through four popular saliency algorithms show that on \textbf{49/56 (87.5\%)} cases - considering both the overall dataset and crucial driving scenarios, SAGE outperforms existing techniques without any additional computational overhead during the training process. The augmented dataset along with the relevant code are available as part of the supplementary material.\footnote{Supplementary material including code and videos are available at~\url{https://sites.google.com/eng.ucsd.edu/sage-net}.} 
\end{abstract}

\input{files/introduction.tex}
\input{files/related_work.tex}
\input{files/methodology.tex}
\input{files/expt_result.tex}
\input{files/conclusion.tex}
\input{files/acknowledgement.tex}
{\small
\bibliographystyle{ieee_fullname}
\bibliography{egpaper_final}
}

\end{document}

%% file: files/introduction.tex
\section{Introduction}

Cameras are one of the most powerful sensors in the world of robotics as they capture detailed information about the environment, and thus can be used for object detection \cite{wang2019salient,liu2010learning} and segmentation \cite{wang2015saliency,wang2017video} - something that is much harder to achieve with a basic range sensor. However, an image/video may contain some irrelevant information. Therefore, there is a need to filter out these unimportant regions and instead, learn to focus our "attention" on parts of the image which are necessary to solve the task at hand. This is  crucial for autonomous driving scenarios, where a vehicle should pay more attention to other vehicles, pedestrians and cyclists present in its vicinity, while ignoring the inconsequential objects. Upon successfully identifying the objects of interest, the controller driving the vehicle only needs to attend to them in order to make optimal decisions.

\input{images/contingency_images.tex}
\input{images/gt_comparisons.tex}
We propose a novel framework for predicting driver's focus of attention through a learnt saliency map by taking into consideration the semantic context in an image. Typical saliency prediction algorithms \cite{palazzi2018predicting, palazzi2017learning, xia2018predicting, tawari2018learning} in driving scenarios rely only on human-gaze information, either through an in-car \cite{Alletto_2016_CVPR_Workshops}, or in-lab \cite{xia2018predicting} setting. However, gaze by itself does not completely describe everything a driver should attend to, mainly due to the following reasons:

(i) \textbf{Peripheral vision:} Humans have a tendency to rely on peripheral vision, thus giving us the ability to fixate our eyes on one object while attending to another. This cannot be captured by an eye-tracking device. Thus, only in-car driver gaze \cite{Alletto_2016_CVPR_Workshops} does not convey sufficient information. While the in-lab annotation does alleviate this problem to some extent \cite{xia2018predicting} by aggregating the gazes of multiple independent observers, it does not completely remove it since that relies on real human gaze too. 

(ii) \textbf{Single focus:} When a human driver realizes that the trajectory of an incoming car or pedestrian is not likely to collide with that of the ego-vehicle, their tendency is to shift the gaze away from the oncoming traffic as it approaches. This is a major cause of accidents. To address that, we propose a method of tracking the motion of every driving-relevant object by detecting it's instances until it goes beyond the field of view of the camera. This is possible because the limitation of a human's ability of single focus does not apply to an autonomous vehicle system. 

(iii) \textbf{Distracted gaze:} A human driver while driving the car might often get distracted by some road-side object - say a brightly colored building, or some attractive billboard advertisement etc. We take care of this issue by only training to detect those objects which influence the task of driving. The in-lab gaze \cite{xia2018predicting} also eliminates this noise by averaging the eye movements of independent observers. However, they assume that the people annotating are positioned in the co-pilot's seat, and therefore cannot realistically emulate a driver's gaze. 

(iv) \textbf{Center-bias:} For majority of a driving task, human gaze remains on the road in front of the vehicle as this is where the vehicle is headed to. When deep learning models are trained on this gaze map, they invariably recognize this pattern and learn to keep the focus there. However, this is not enough since there might be important regions away from the center of the road which demand attention - such as when cars or pedestrians approach from the sides. Thus, relying only gaze data does not help capture these important cues.

Figure \ref{fig:contingency_imgs} shows an example of an accident-prone situation, where the predicted saliency maps from an algorithm trained using different target labels are shown. Gaze-only models were able to detect the car ahead, but completely missed the pedestrian jaywalking. In contrast, our approach successfully detects both objects since it has learnt to predict semantic context in an image.

It is important to note, however, that semantics alone does not completely provide insights into the action that a driver might take at run-time. This is because a saliency map obtained only from training on semantics will give an equal-weighted attention on all the objects present. Also, when there is no object of relevance (\ie an empty road near the countryside), this saliency map will not provide any attention. In reality, here the focus should be towards road boundaries, lane dividers, curbs etc. These regions can be effectively learnt through gaze information which is an indicator of a driver's intent. Thus, we design a Semantics Augmented GazE (SAGE) \textit{ground-truth}, which successfully captures both gaze and semantic context. Figure \ref{fig:gt-comparison} shows how our proposed ground-truth looks as compared to the existing gaze-only ground-truths.

There are three novel contributions made in this paper. Firstly, we propose SAGE - a combined attention mechanism, that can be used to train saliency models for accurately predicting an autonomous vehicle's (hereafter termed as driver) focus of attention. Secondly, we provide a thorough saliency detection framework - SAGE-Net, by including important cues in driving such as distance to objects (depth), speed of ego-vehicle and pedestrian crossing intent to further enhance the initial raw prediction obtained from SAGE. Finally, we conduct a series of experiments using multiple saliency algorithms on different driving datasets to evaluate the flexibility, robustness, and adaptability of SAGE - both over the entire dataset, and also specific important driving scenarios such as intersections and busy traffic regions. The rest of the paper is organized as follows. Section \ref{sect:related work} discusses the existing state-of-the-art research in driver saliency prediction. Section \ref{sect:methodology} then provides details of the proposed framework, followed by the extensive experiments conducted in Section \ref{sect:expt_res}. Finally, Section \ref{sect:conclusion} concludes the discussion and mentions the real-world implication of the conducted research.

%% file: images/contingency_images.tex
\begin{figure}[!t] %
    \centering
    \subfloat[Input image] {{\includegraphics[width=3.7cm]{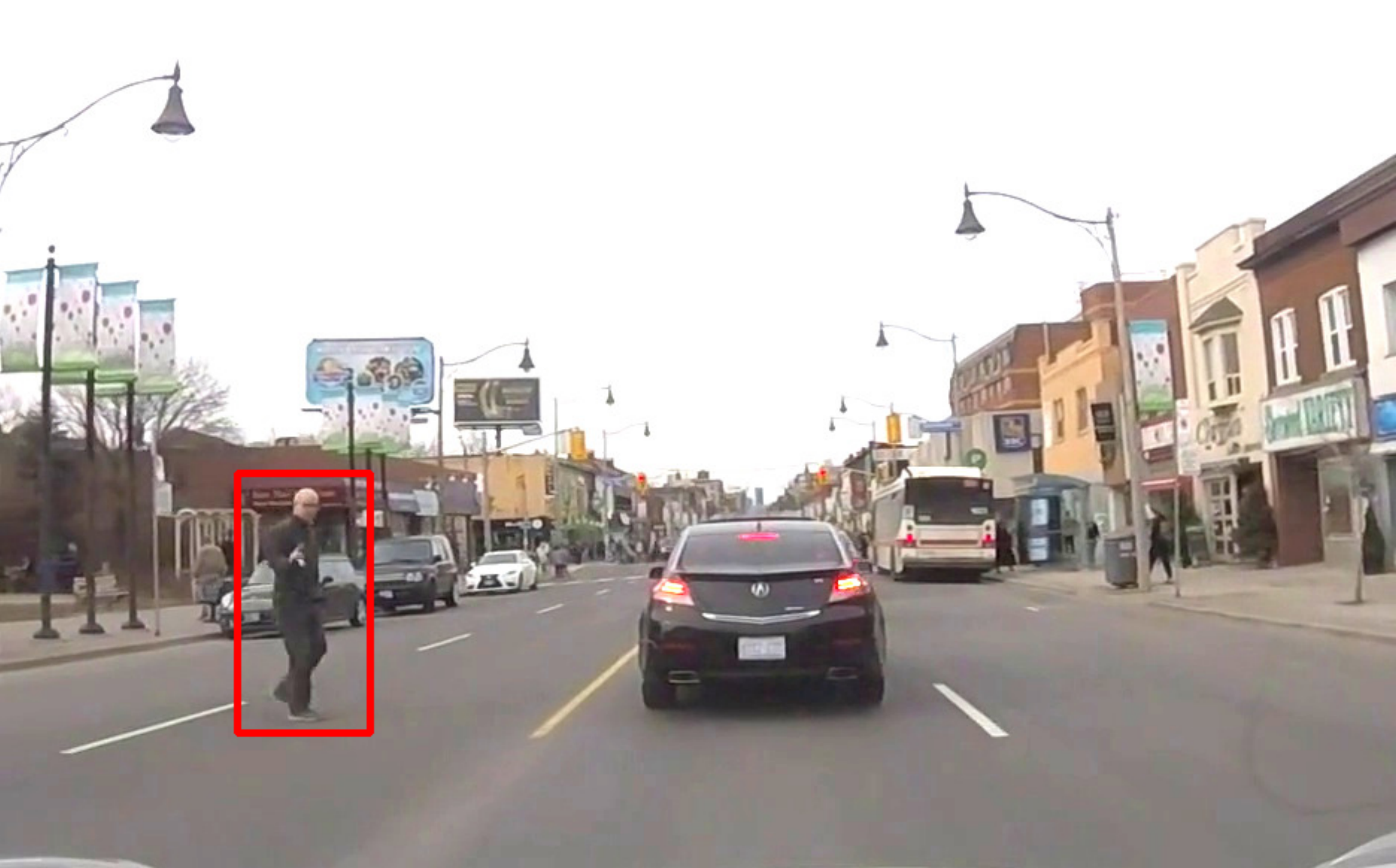} }}%
    \quad
    \subfloat[\textbf{SAGE-Net (our)}] {{\includegraphics[width=3.7cm]{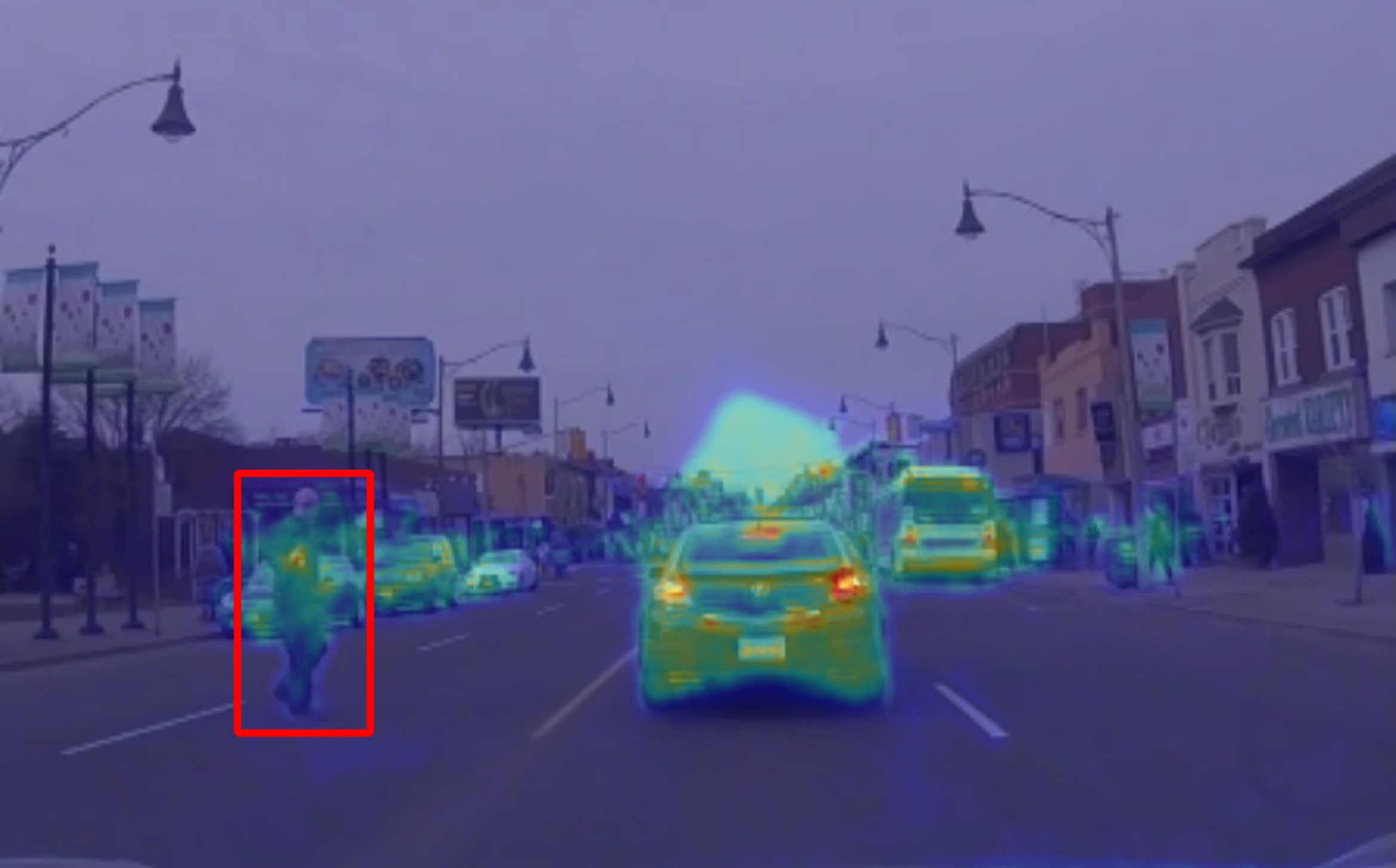} }}%
    \quad
    \\
    \subfloat[BDD-A \cite{xia2018predicting}] {{\includegraphics[width=3.7cm]{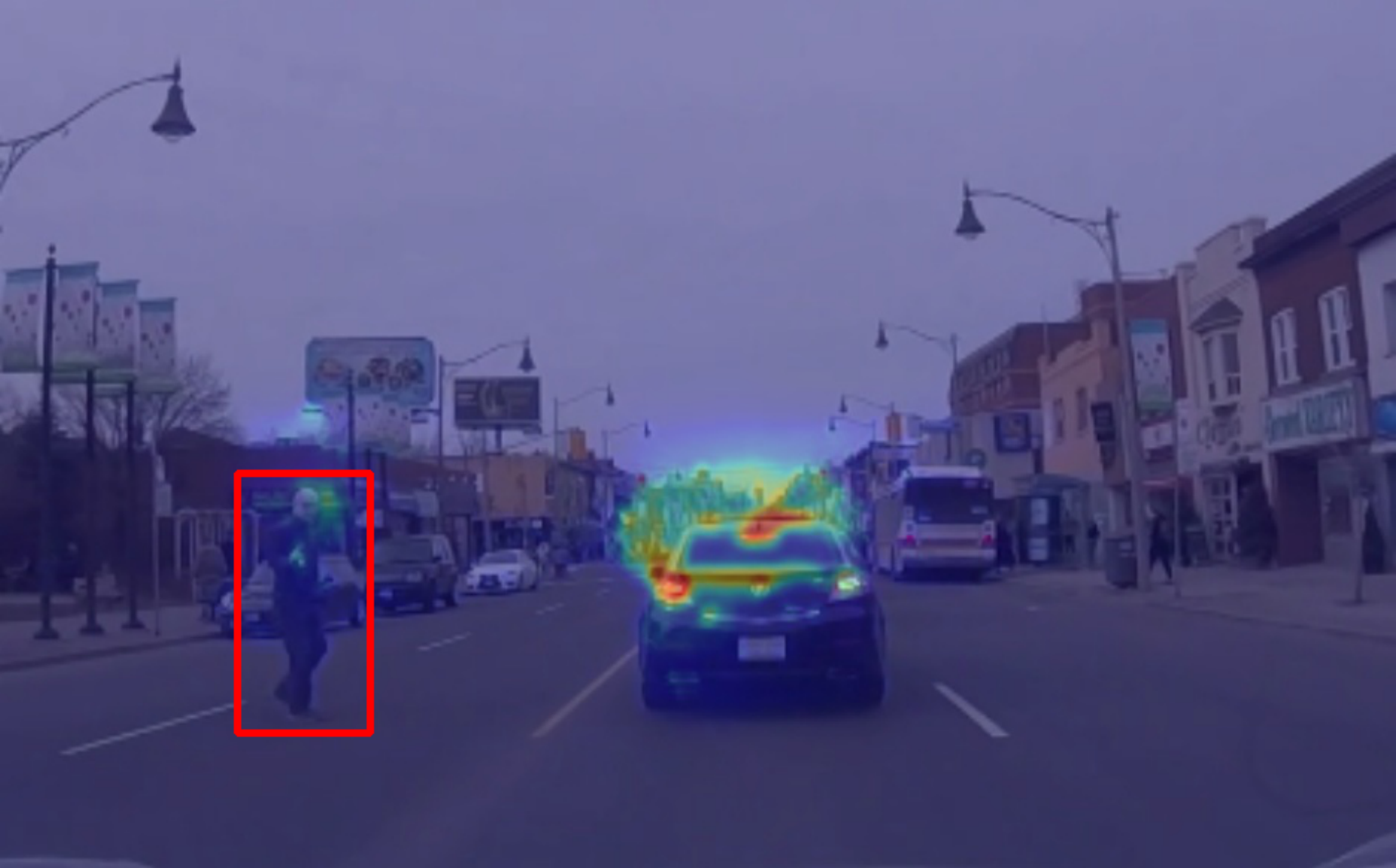} }}%
    \quad
    \subfloat[DR(eye)VE \cite{Alletto_2016_CVPR_Workshops}] {{\includegraphics[width=3.7cm]{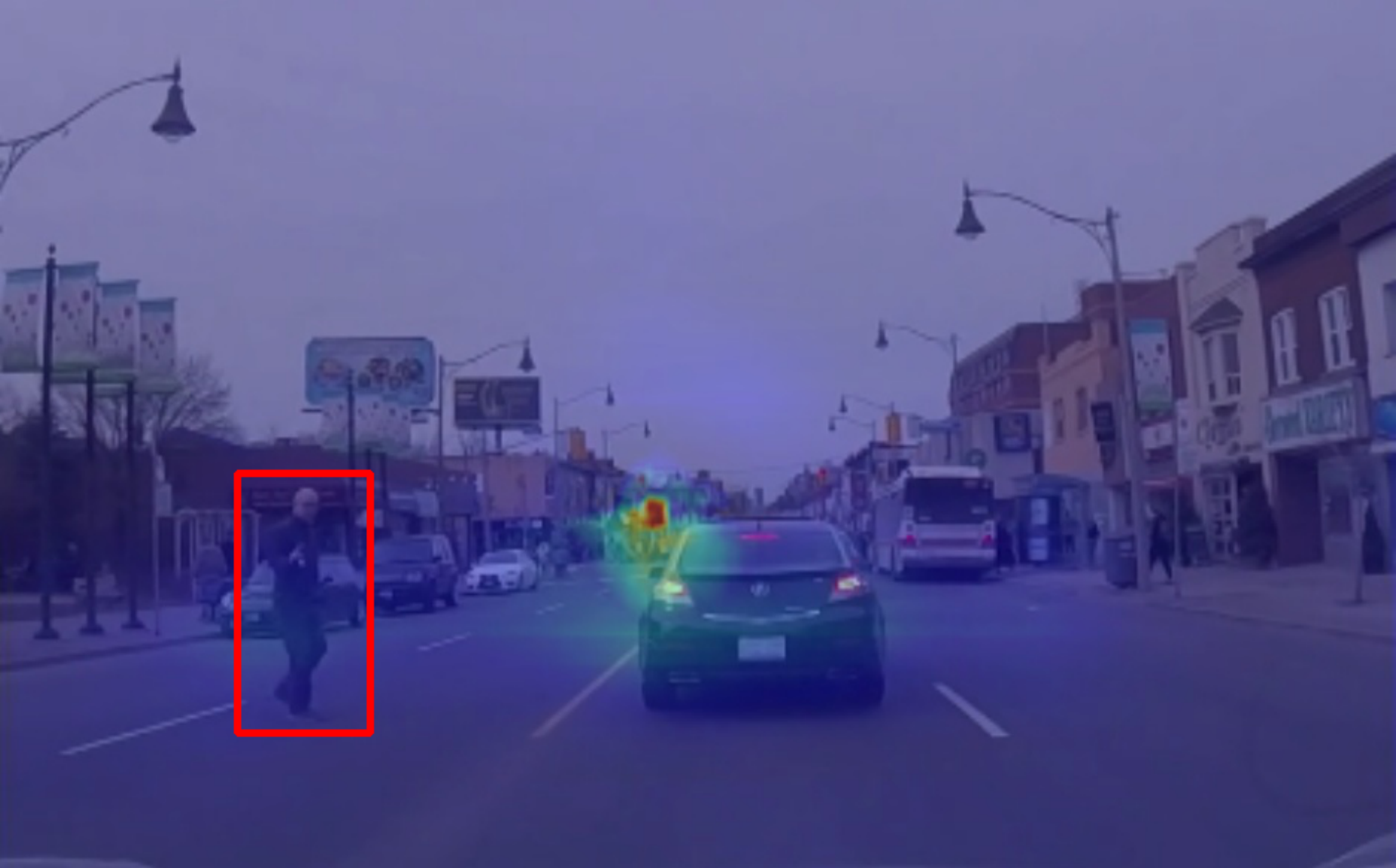} }}%
    \quad
    \caption{Predicted saliency map for different models (Best viewed in color). The bounding box shows a pedestrian illegally crossing the road and is prone to accident. While other models only capture the car ahead (partially), our proposed model can \textbf{completely} learn to detect both the car and the crossing pedestrian.}%
    \label{fig:contingency_imgs}%
\end{figure}

%% file: images/gt_comparisons.tex
\begin{figure*}[!b] %
    \centering
    \subfloat[RGB image 1]{{\includegraphics[width=4.5cm]{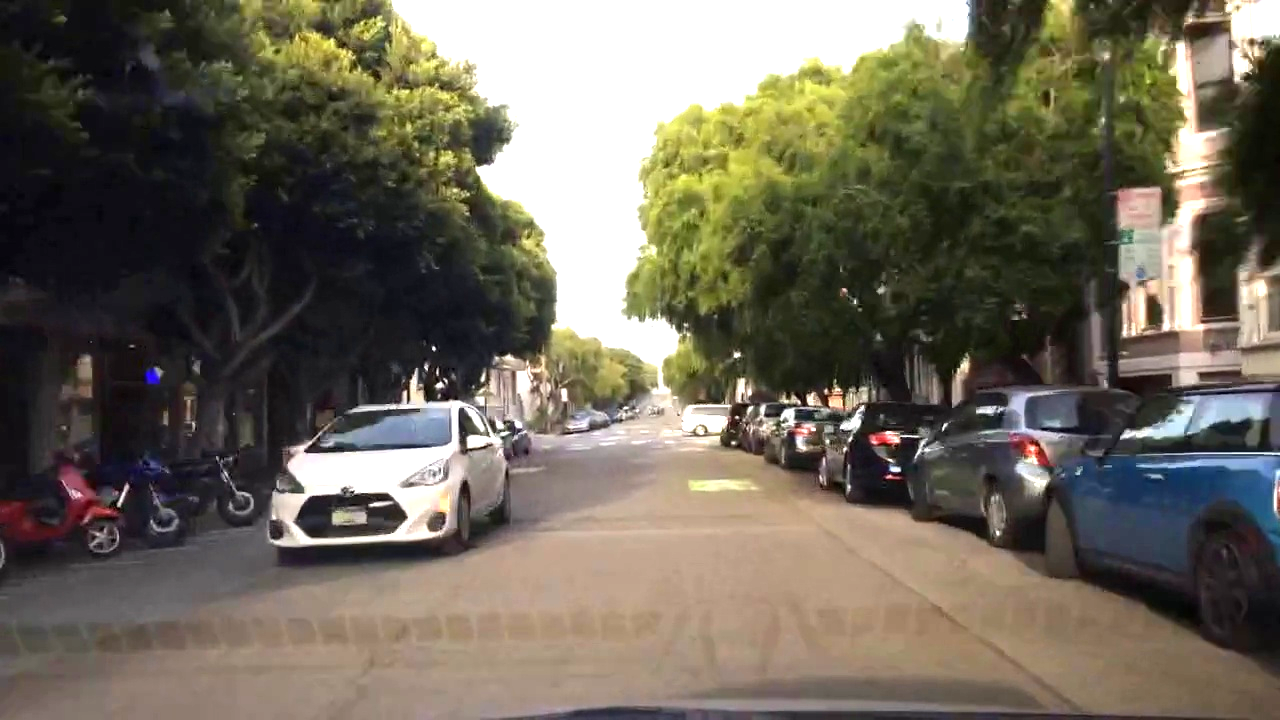} }}%
    \quad
    \subfloat[Gaze-only groundtruth]{{\includegraphics[width=4.5cm]{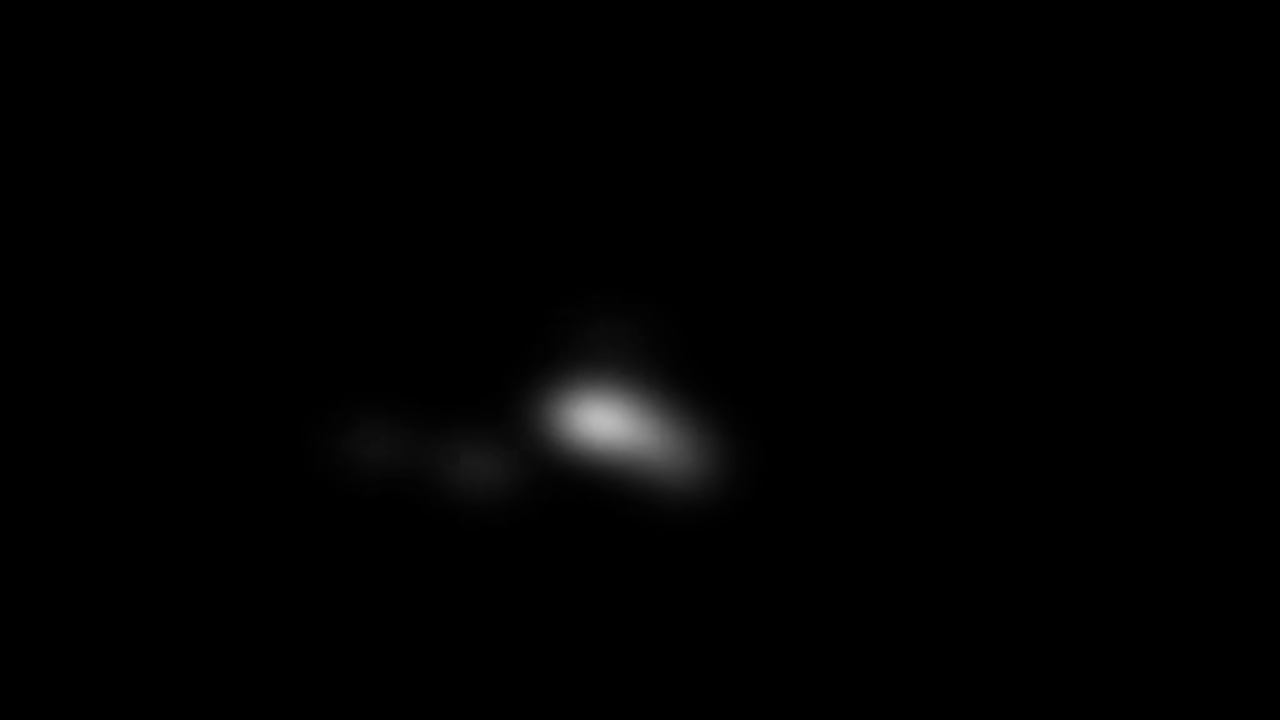} }}%
    \quad
    \subfloat[SAGE groundtruth (ours)]{{\includegraphics[width=4.5cm]{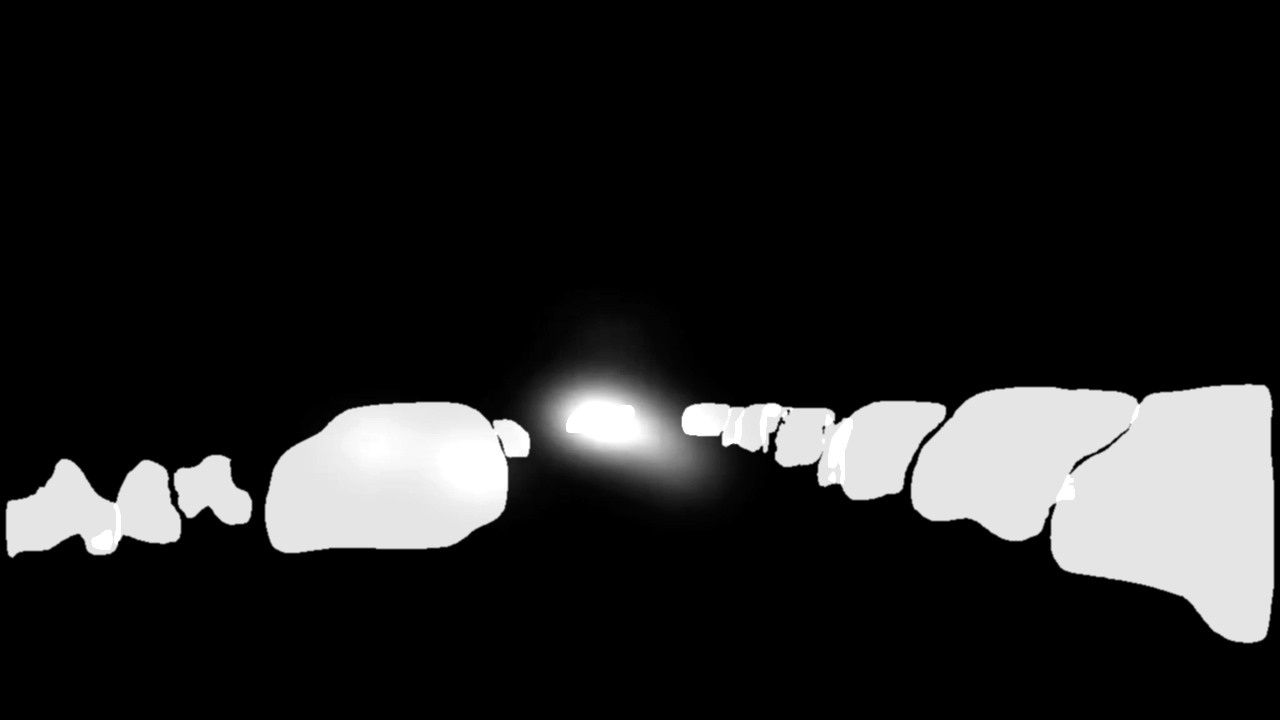} }}%
    \\
    \subfloat[RGB image 2]{{\includegraphics[width=4.5cm]{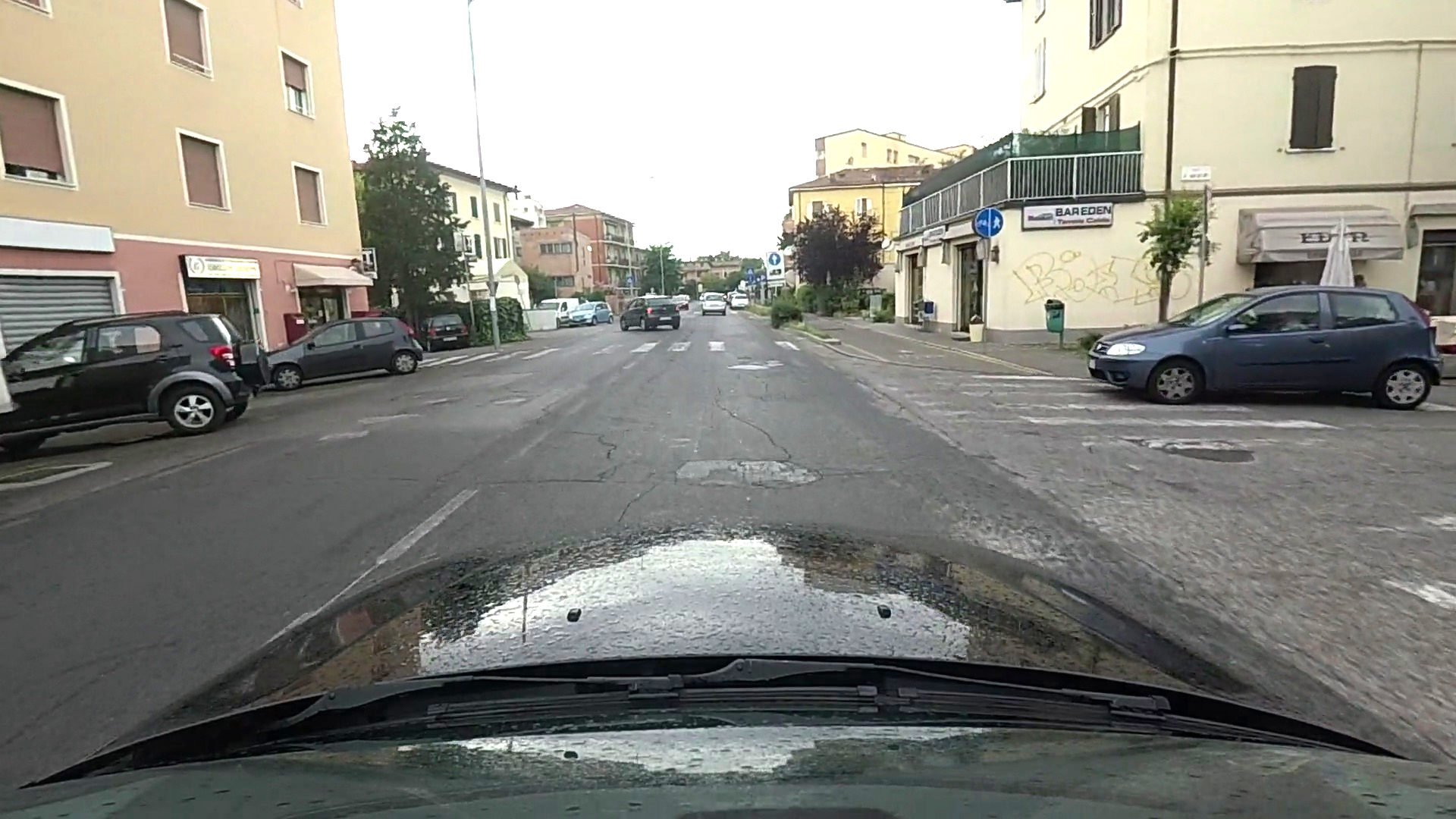} }}%
    \quad
    \subfloat[Gaze-only groundtruth]{{\includegraphics[width=4.5cm]{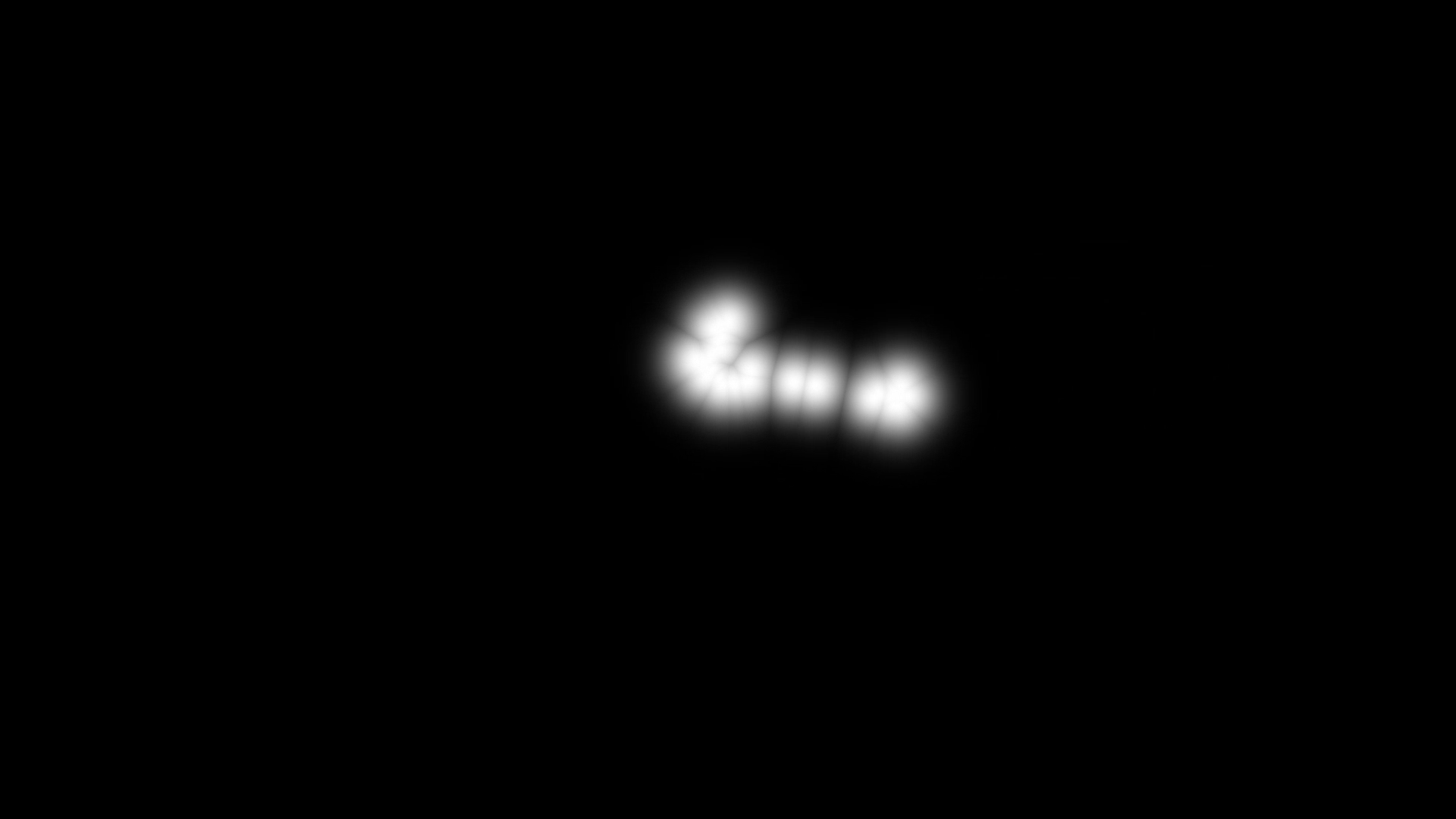} }}%
    \quad
    \subfloat[SAGE groundtruth (ours)]{{\includegraphics[width=4.5cm]{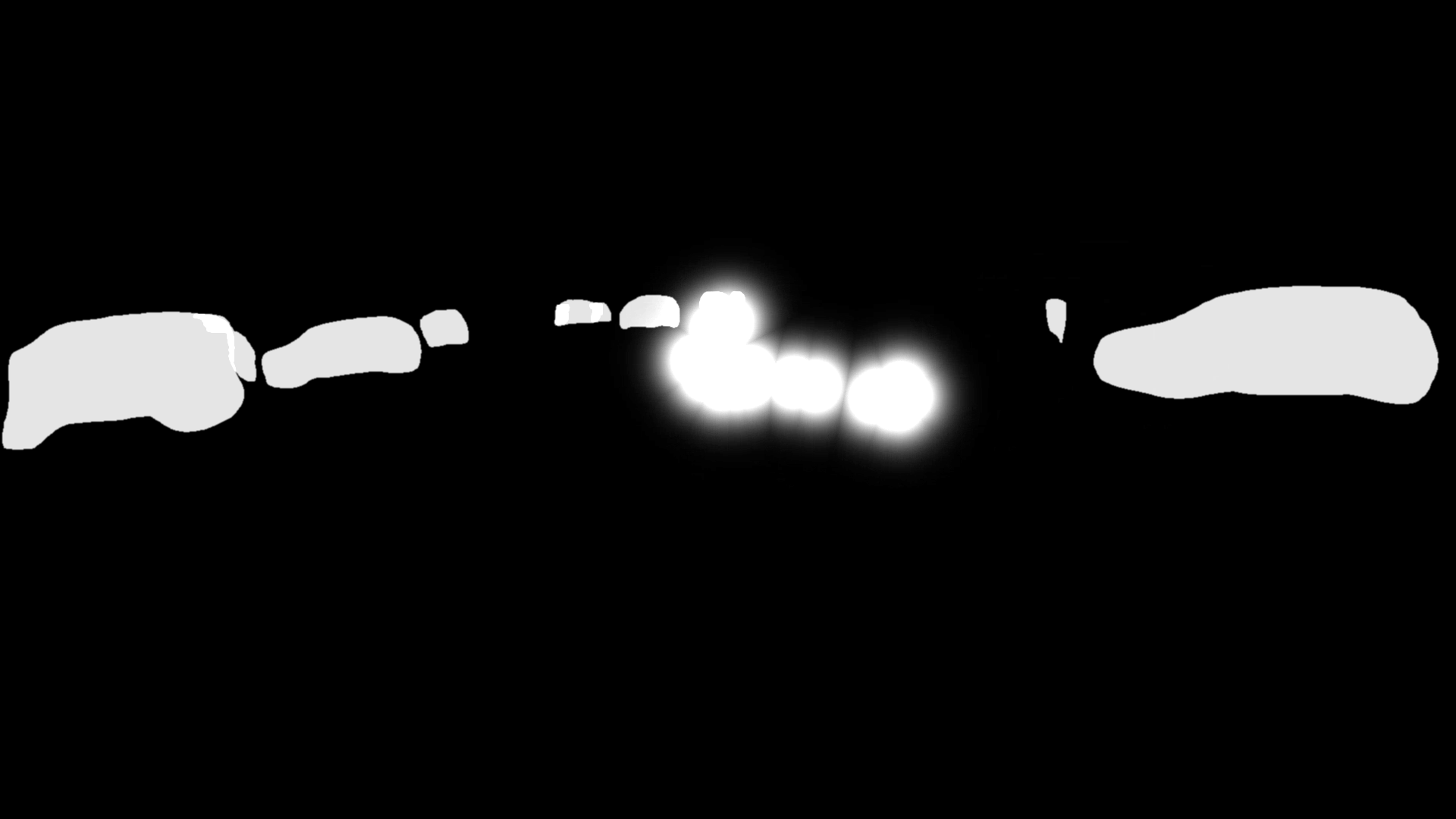} }}%
    \caption{Comparison of SAGE with the existing gaze-only groundtruths. The top row [a-c] is for the BDD-A dataset \cite{xia2018predicting} while the bottom row [d-f] is for the DR(eye)VE dataset \cite{Alletto_2016_CVPR_Workshops}. The gaze-only maps indicate the heading of the ego-vehicle, but completely ignore the nearby and incoming cars. In contrast, SAGE captures both the driver's intent and the relevant objects.}
    \label{fig:gt-comparison}%
\end{figure*}

%% file: files/related_work.tex
\section{Related Work} \label{sect:related work}

\textbf{Advances in Salient Object Detection:} Detection \cite{wang2019salient, liu2010learning} and segmentation \cite{wang2015saliency,wang2017video} of salient objects in the natural scene has been a very active area of research in the computer vision community for a long time. One of the earliest works in saliency prediction, by Itti \etal \cite{itti1998model}, considered general computational frameworks and psychological theories of bottom-up attention, based on center-surround mechanisms \cite{treisman1980feature, wolfe1989guided, koch1987shifts}. Subsequent behavioral \cite{parkhurst2002modeling} and computational investigations \cite{bruce2006saliency} used "fixations" as a means to verify the saliency hypothesis and compare models. Our approach differs from them as we incorporate both a bottom-up strategy by scanning through the entire image and detecting object features that are relevant for driving, as well as a top-down strategy by incorporating human gaze which is purely task driven. Some later studies \cite{liu2010learning, achanta2008salient} defined saliency detection as a binary segmentation problem. We adopt a similar strategy, but instead of using handcrafted features that do not generalize well to real world scenes, we use deep learning techniques for robust feature extraction. Since the introduction of Convolutional Neural Networks (CNNs), a number of approaches have been developed for learning global and local features through varying receptive fields, both for 2D image datasets \cite{wang2019salient, liu2019end, choe2019attention, fu2019dual}, and video-based saliency predictions \cite{wang2019learning, lu2019see, fan2019shifting, 8968108}. However, these algorithms are either too heavily biased towards image datasets, or involve designs of complicated architectures which make them difficult to train. In contrast, our approach helps to improve existing architectures without any additional training parameters, thereby keeping the complexity unchanged. This is very important for an autonomous system since we want to make it as close to real-time as possible. For a detailed survey of salient object detection, we refer the reader to the work by Borji \etal \cite{borji2014salient}.

\textbf{Saliency for driving scenario:} Lately, there has been some focus on driver saliency prediction due to rise of the number of driving \cite{kotseruba2016joint, yu2018bdd100k, Ramanishka_behavior_CVPR_2018, 360LiDARTracking_ICRA_2019, narayanan2019dynamic} and pedestrian tracking \cite{Dollar2012PAMI, ess2007depth, kotseruba2016joint} datasets. Most saliency prediction models are trained using human gaze information, either through in-car eye trackers \cite{Alletto_2016_CVPR_Workshops, palazzi2018predicting}, or through in-lab simulations \cite{xia2018predicting, tawari2018learning}. However, as discussed above, these methods only give an estimate of the gaze, which is often prone to center bias, or distracted focus. In contrast, our approach involves combining scene semantics along with the existing gaze data. This ensures that the predicted saliency map can effectively mimic a real driver's intent, with the added feature of also being able to successfully detect and track important objects in the vicinity of the ego-vehicle.

%% file: files/methodology.tex
\section{SAGE-Net: Semantic Augmented GazE detection Network} \label{sect:methodology}

Figure \ref{full_architecture} provides a simplified illustration of the entire SAGE-Net framework, which comprises of three components: a SAGE detection module, a distance-based attention update module, and finally a pedestrian intent-guided saliency module. We begin by firstly describing how the SAGE maps are obtained in \S\ref{sage}. Next, in \S\ref{depth}, we describe how relative distances of objects from ego-vehicle should impact saliency prediction. Lastly, in \S\ref{ped}, we highlight the importance of pedestrian crossing intent detection and how it influences the focus of attention.

\subsection{SAGE saliency map computation} \label{sage}
\input{images/full_architecture.tex}

We propose a new approach to predicting driving attention maps which not only uses raw human gaze information, but also learns to detect the scene semantics directly. This is done using the Mask R-CNN (M-RCNN) \cite{he2017mask} object detection algorithm, which returns a segmented mask around an object of interest along with it's identity and location. 

We used the Matterport implementation of M-RCNN \cite{matterport_maskrcnn_2017} which is based on Feature Pyramid Network (FPN) \cite{lin2017feature} and uses ResNet-101 \cite{he2016deep} as backbone. The model is trained on the MS-COCO dataset \cite{lin2014microsoft}. However, out of the total 80 objects in \cite{lin2014microsoft}, we select 12 categories which are most relevant to driving scenarios - $\mathtt{person}$, $\mathtt{bicycle}$, $\mathtt{car}$, $\mathtt{motorcycle}$, $\mathtt{bus}$, $\mathtt{truck}$, $\mathtt{traffic}$ $\mathtt{light}$, $\mathtt{fire}$ $\mathtt{hydrant}$, $\mathtt{stop}$ $\mathtt{sign}$, $\mathtt{parking}$ $\mathtt{meter}$, $\mathtt{bench}$ and $\mathtt{background}$. For each video frame, M-RCNN provides an instance segmentation of every detected object. However, as the relative importance of different instances of the same object is not a significant cue, we stick to a binary classification approach where we segment all objects vs the background. This object-level segmented map is then superimposed on top of the existing gaze map provided by a dataset, so as to preserve the gaze information. This gives us the final saliency map as seen in Fig \ref{fig:gt-comparison}. Upon inspection, it can be clearly seen that our ground-truth has managed to capture a lot more semantic context from the scene, which gaze-only maps have missed.

\subsection{Does relative distance between objects and ego-vehicle impact focus of attention?} \label{depth}
Depth estimation through supervised \cite{eigen2014depth, liu2015learning, mayer2016large} and unsupervised \cite{garg2016unsupervised, vijayanarasimhan2017sfm} learning methods as a measure of relative distance between objects and ego-vehicle has been a long studied problem in the autonomous driving community \cite{mahjourian2018unsupervised, godard2017unsupervised, godard2018digging}. Human beings inherently react and give more attention to vehicles and pedestrians which are "closer" to them as opposed to those at a distance, since chances of collision are much higher for the former case. Unfortunately, this crucial information is yet to be exploited for predicting driving saliency maps to the best of our knowledge. In this paper, we consider this through the recently proposed self-supervised monocular depth estimation approach - Monodepth2 \cite{godard2018digging}. However, SAGE-Net is not restricted to just this algorithm, but can effectively inherit stereo or LiDAR-based depth estimators into its framework as well.

We considered two methods of incorporating depth maps into our framework. The first involves taking a parallel depth channel which does not undergo any training, but is simply used to amplify nearby regions of the predicted saliency map. The second method is to use it as a separate trainable input to the saliency prediction model along with the raw image, in a manner similar to how optical flow and semantic segmentation maps are trained in \cite{palazzi2018predicting}. We decided to go with the first strategy because in addition to being much simpler and faster to implement, it also removes the issue of training a network only on depth map which has a lot less variance in data, thus leading to overfitting towards the vanishing point in the image. 

Given an input clip of 16 image frames, $\mathbf{\mathcal{X}_{RGB}} \in \mathbb{R}^{16\times 3\times h\times w}$, we obtain the raw prediction $\mathcal{Y}_{\text{RGB}} \in \mathbb{R}^{h\times w}$. In addition, for each frame, we also compute the depth map $\mathcal{D}_{\text{RGB}} \in \mathbb{R}^{h\times w}$. Finally, we combine the raw prediction with the depth map to obtain $\mathcal{Y}_{\text{RGB-D}}$ using the $\oplus$ operator, which is defined as 
\begin{equation}
    \mathcal{Y}_{\text{RGB}} \oplus \mathcal{D}_{\text{RGB}} = \mathcal{Y}_{\text{RGB}}*\mathcal{D}_{\text{RGB}} + \mathcal{Y}_{\text{RGB}}
\end{equation}
\subsection{Should we pay extra attention to pedestrians crossing at intersection scenarios?} \label{ped}
Accurate pedestrian detection in crosswalks is a vital task for an autonomous vehicle. Thus, we include an additional module which focuses solely on the crossing intent of pedestrians at intersections, and correspondingly updates the saliency prediction. It should be noted that even though SAGE does capture pedestrians in its raw prediction in general driving scenarios, it does not distinguish between them and other objects in crowded traffic conditions such as intersections. This is critical since the chances of colliding with a pedestrian are much higher around intersection regions than at other roads. However, this is a slow process since it involves detecting pedestrians and predicting their pose at run-time. Fortunately, this situation only occurs when the speed of the ego-vehicle itself is less. Thus, we only include this in our framework when the speed of the ego-vehicle ($\mathtt{v_{ego}}$) is below a certain threshold velocity $\mathtt{v_{thresh}}$. It is not very difficult to obtain $\mathtt{v_{ego}}$ since most driving datasets provide this annotation \cite{Alletto_2016_CVPR_Workshops, xia2018predicting}. Also, for an autonomous vehicle, the odometry reading contains this. $\mathtt{v_{thresh}}$ is a tunable hyper-parameter which can vary as per the road and weather conditions. When $\mathtt{v_{ego}} < \mathtt{v_{thresh}}$, we look to see if there are pedestrians crossing the road. This is done using the recently proposed algorithm ResEnDec \cite{gujjar:icra19} which predicts the intent $\mathcal{I}$ of pedestrians as "$\mathtt{crossing}$" or "$\mathtt{not}$ $\mathtt{crossing}$" through an encoder-decoder framework using a spatio-temporal neural network and ConvLSTM. We trained this algorithm on the JAAD \cite{kotseruba2016joint} dataset, considering 16 consecutive frames to be the temporal strip while making a prediction on the last frame $\mathcal{X}_{last}$. Our framework is designed such that if the prediction is "$\mathtt{crossing}$", we use an object detector $\mathcal{O}$ such as YOLOv3 \cite{redmon2018yolov3} to get the bounding box of the pedestrians from that last frame. Consequently, we amplify the predicted attention for pixels inside the bounding boxes, while leaving the rest of the image intact. This is given by the $\otimes$ operator, defined as follows
\begin{equation}
    \mathcal{Y}_{\text{RGB-D}} \text{ $\otimes$ } \mathbf{bbox} = \begin{cases} \mathcal{Y}_{\text{RGB-D}}[x,y]*k \text{ $\forall$ } (x,y) \in \mathbf{bbox}\\ \mathcal{Y}_{\text{RGB-D}}[x,y]*1/k \text{, else}\end{cases}
\end{equation}
where $k$ is an amplification factor $(>1)$ by which the predicted map is strengthened. If the predicted intent is "$\mathtt{not}$ $\mathtt{crossing}$", we simply stick with the original prediction $\mathcal{Y}_{\text{RGB-D}}$. The summary of the entire SAGE-Net algorithm is depicted in \ref{algorithm}.
\input{images/models_comparison.tex}
\input{files/algorithm.tex}

%% file: images/full_architecture.tex
\begin{figure*}[t]
  \centering
  \includegraphics[width=\linewidth]{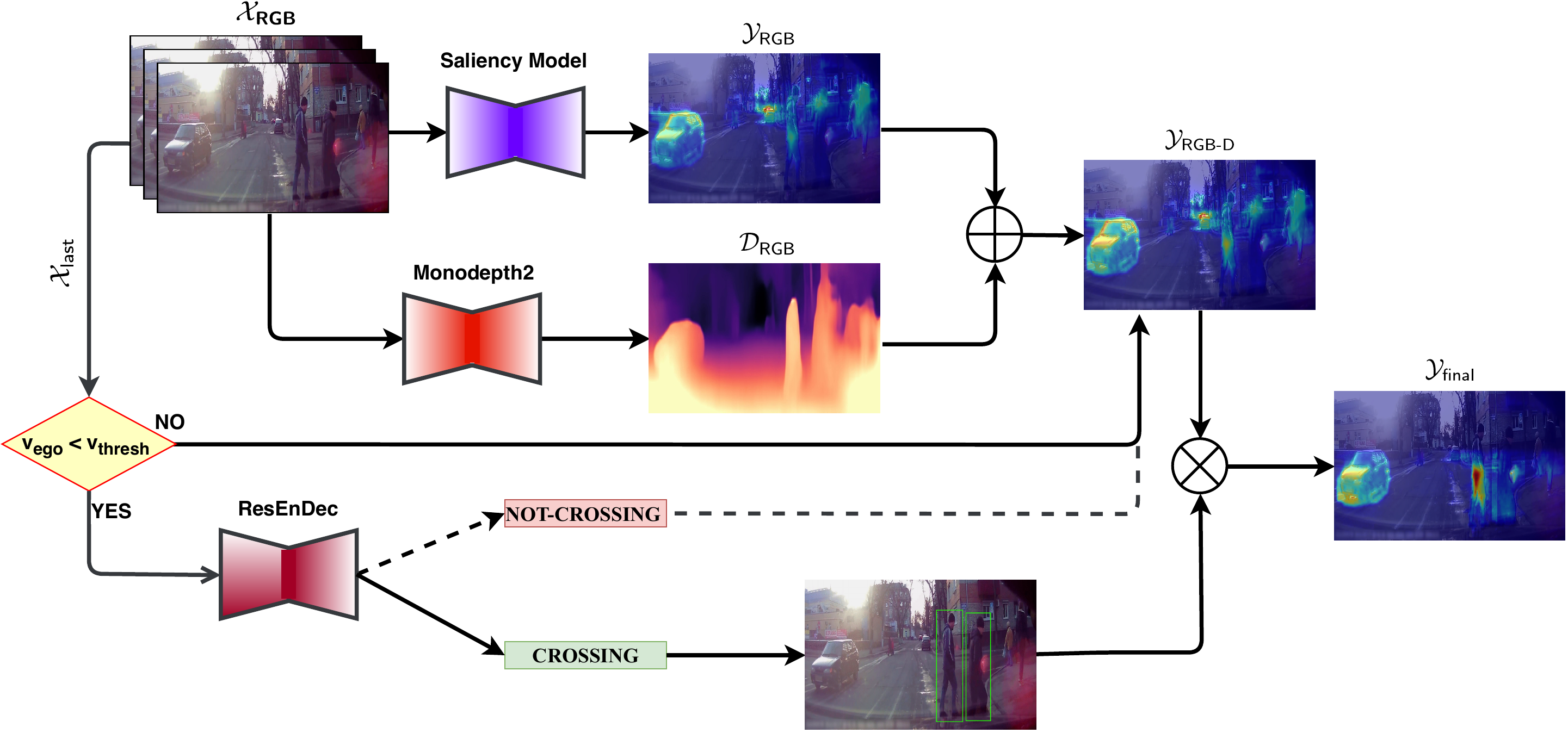}
  \caption{\small The complete \textbf{SAGE-Net} framework (Best viewed in color), comprising of a saliency model trained on \textbf{SAGE} groundtruth, and added parallel modules for depth estimation and pedestrian intent prediction based on ego-vehicle speed ($v_{ego}$).}
  \label{full_architecture}
\end{figure*}

%% file: images/models_comparison.tex
\begin{figure*}[t] %
    \centering
    \subfloat[RGB Image] {{\includegraphics[width=4cm]{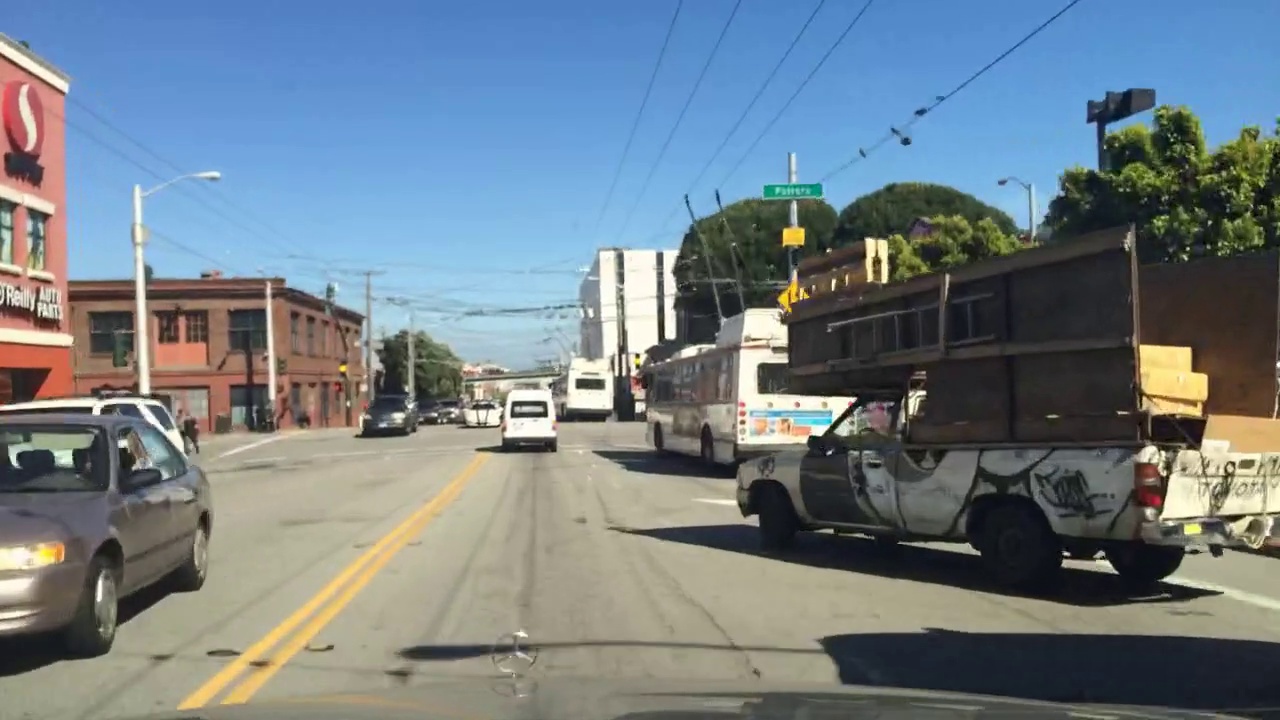} }}%
    \quad
    \\
    \subfloat[DR(eye)VE \cite{palazzi2017learning} with BDDA gt] {{\includegraphics[width=4cm]{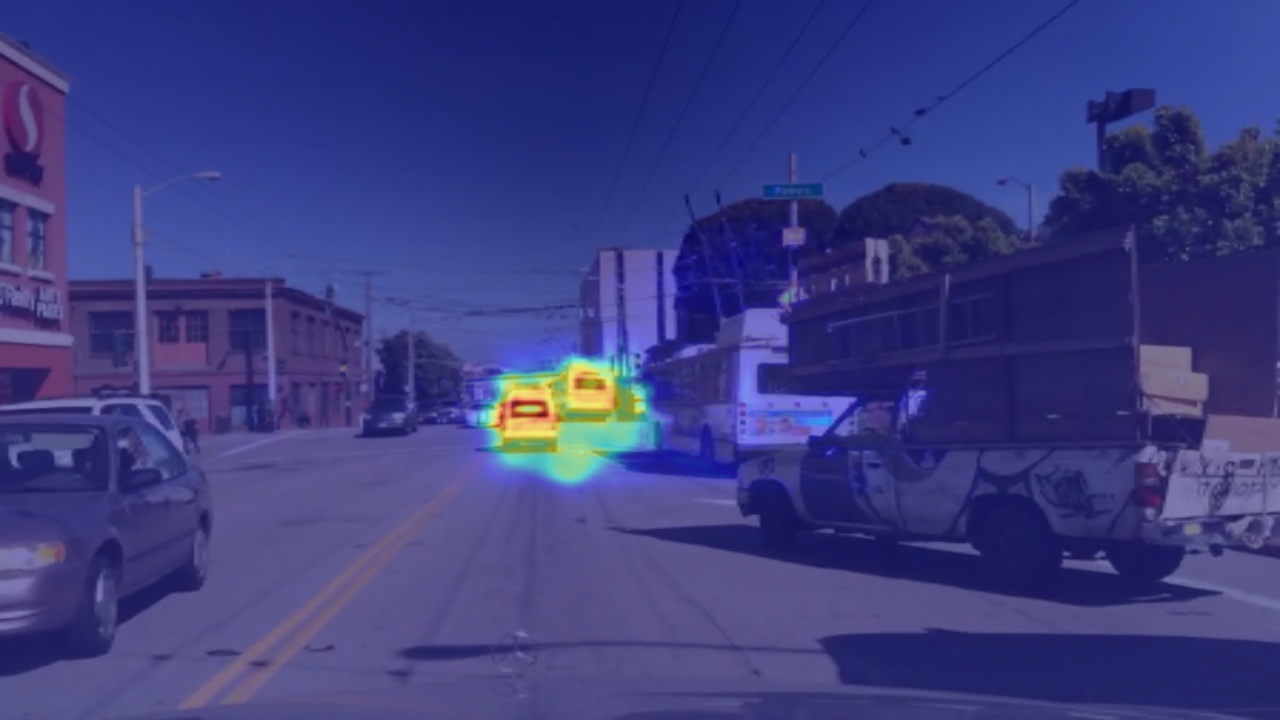} }}%
    \quad
    \subfloat[BDDA \cite{xia2018predicting} with BDDA gt] {{\includegraphics[width=4cm]{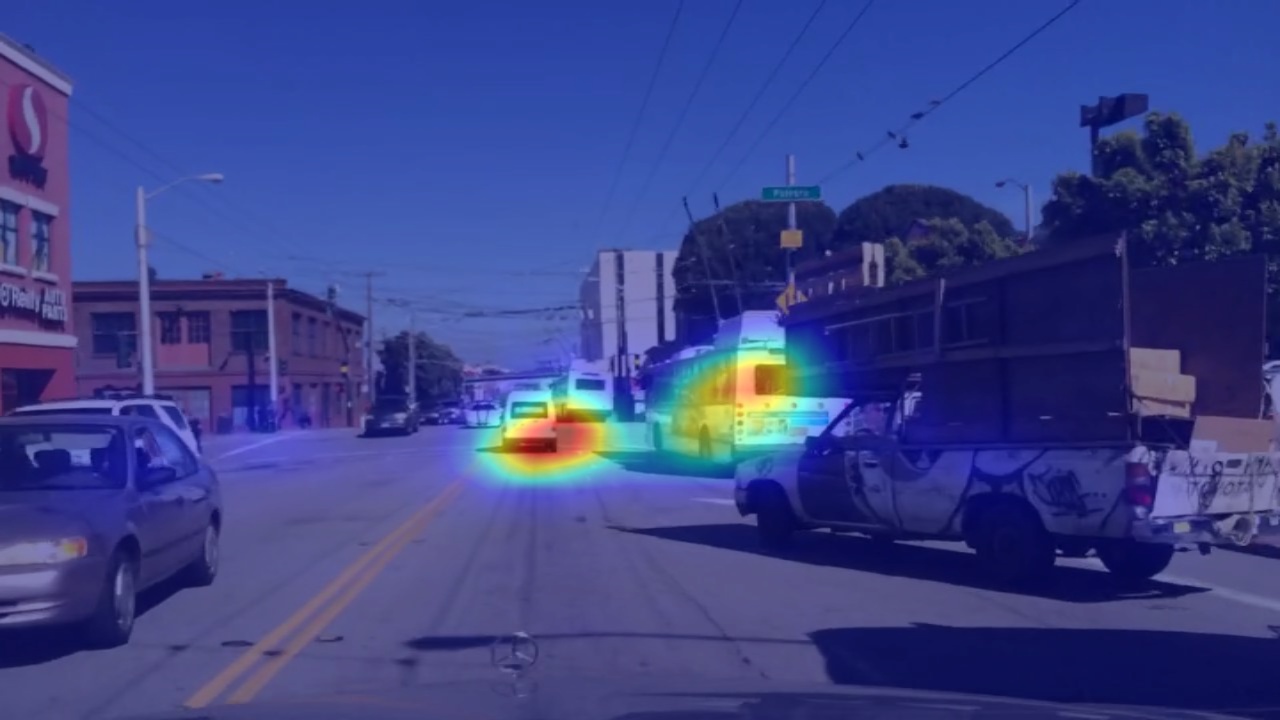} }}%
    \quad
    \subfloat[ML-Net \cite{mlnet2016} with BDDA gt] {{\includegraphics[width=4cm]{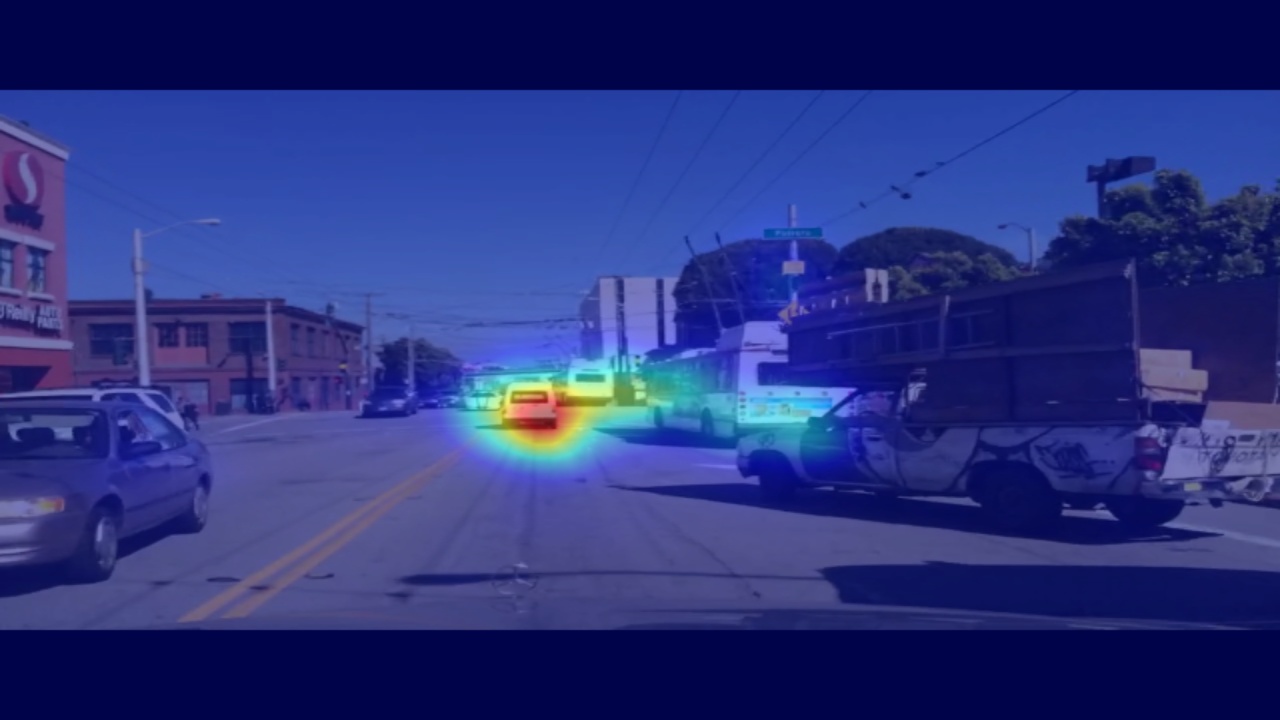} }}%
    \quad
    \subfloat[PiCANet \cite{liu2018picanet} with BDDA gt] {{\includegraphics[width=4cm]{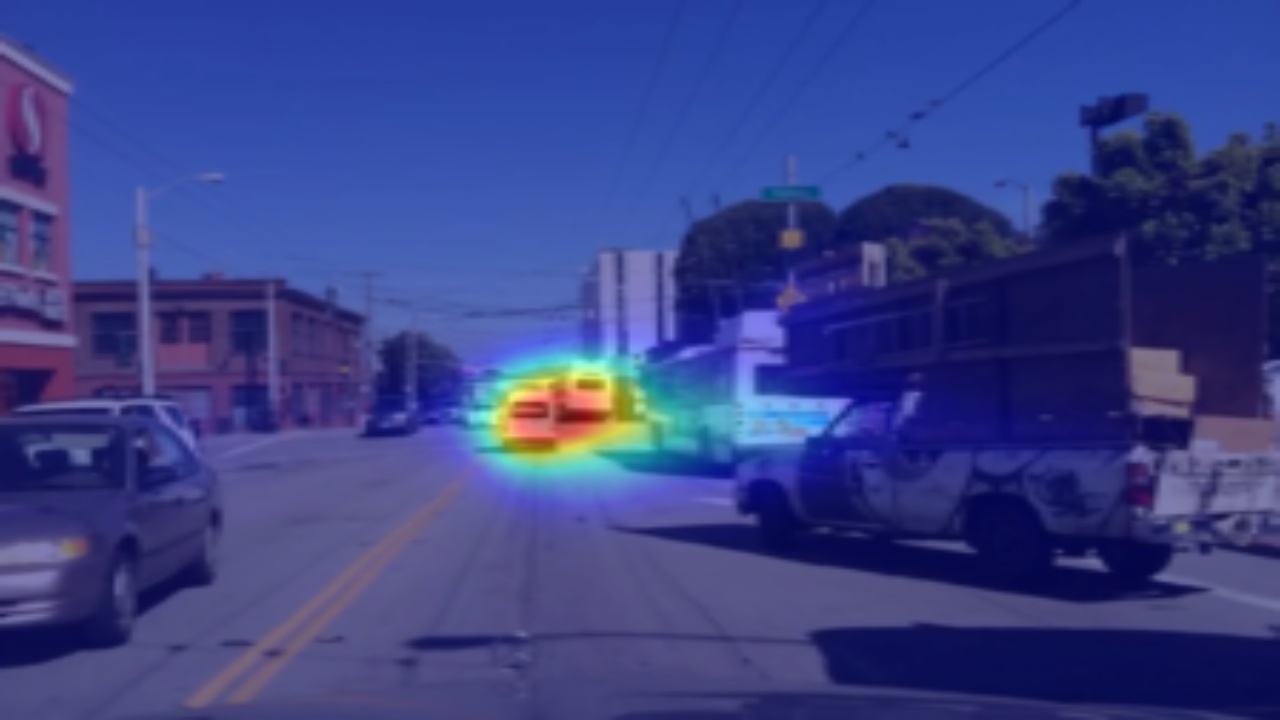} }}%
    \quad
\\
    \subfloat[DR(eye)VE \cite{palazzi2017learning} with SAGE gt] {{\includegraphics[width=4cm]{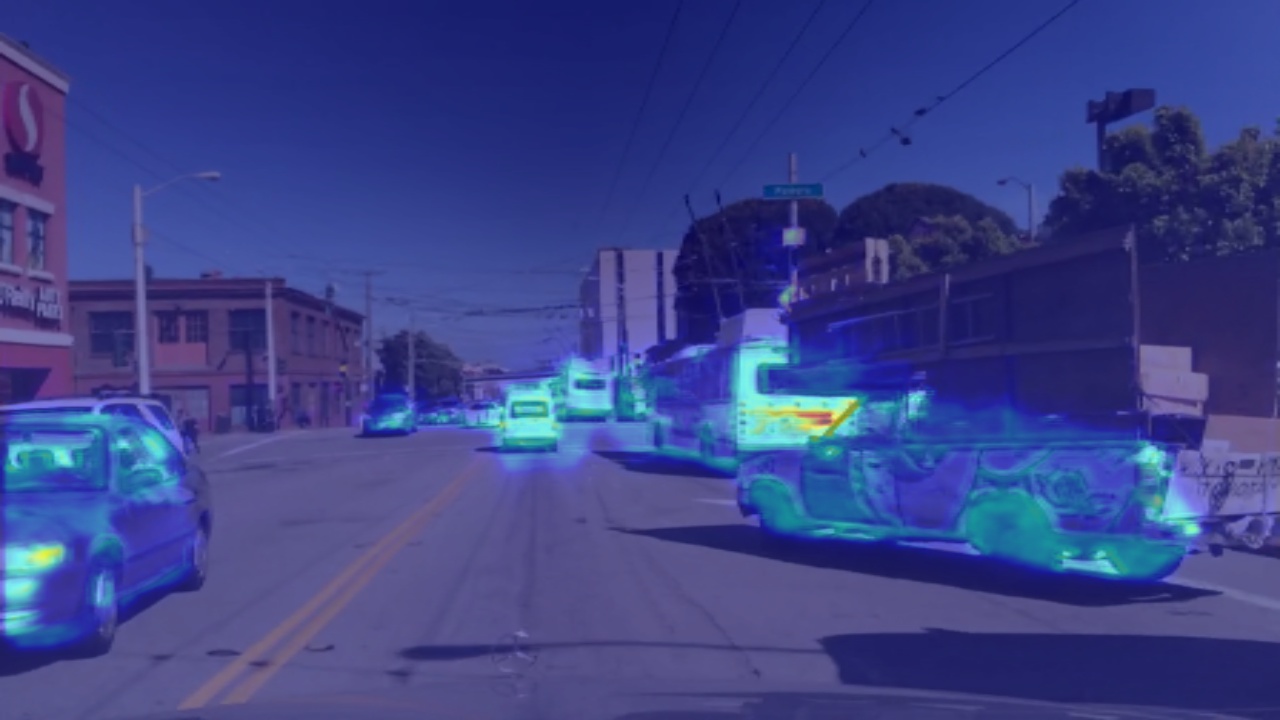} }}%
    \quad
    \subfloat[BDDA \cite{xia2018predicting} with SAGE gt] {{\includegraphics[width=4cm]{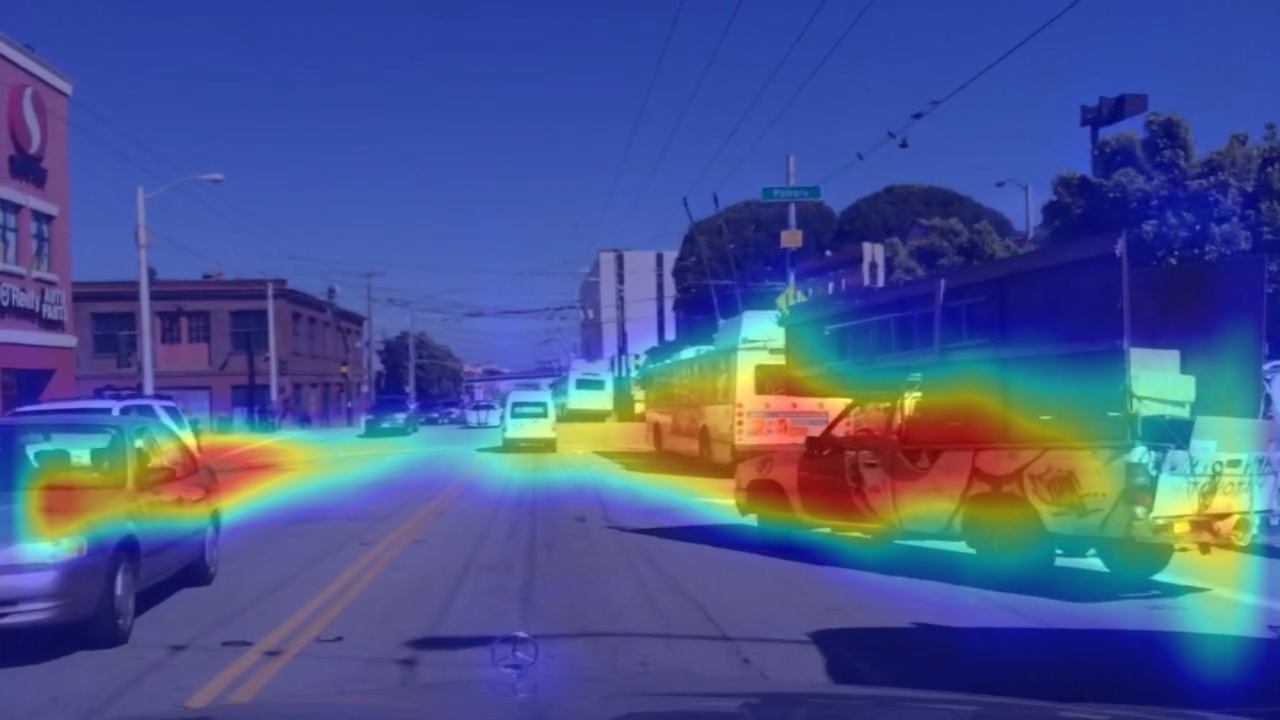} }}%
    \quad
    \subfloat[ML-Net \cite{mlnet2016} with SAGE gt] {{\includegraphics[width=4cm]{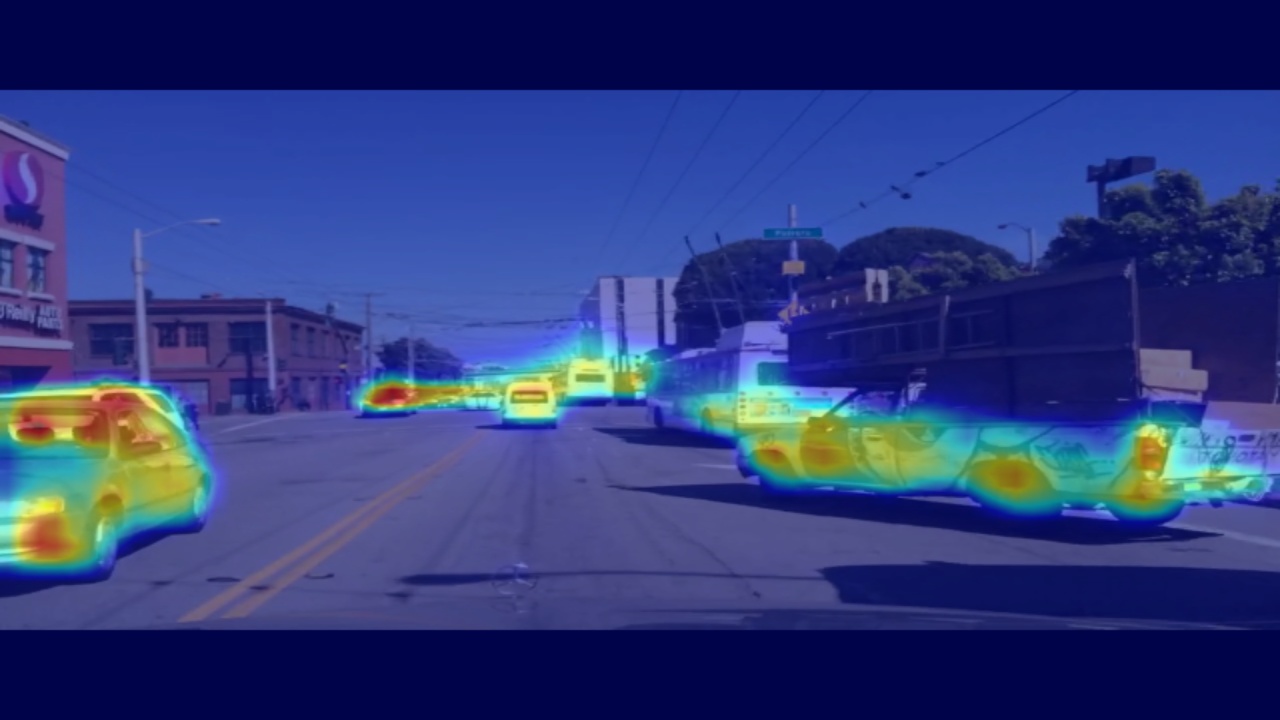} }}%
    \quad
    \subfloat[PiCANet \cite{liu2018picanet} with SAGE gt] {{\includegraphics[width=4cm]{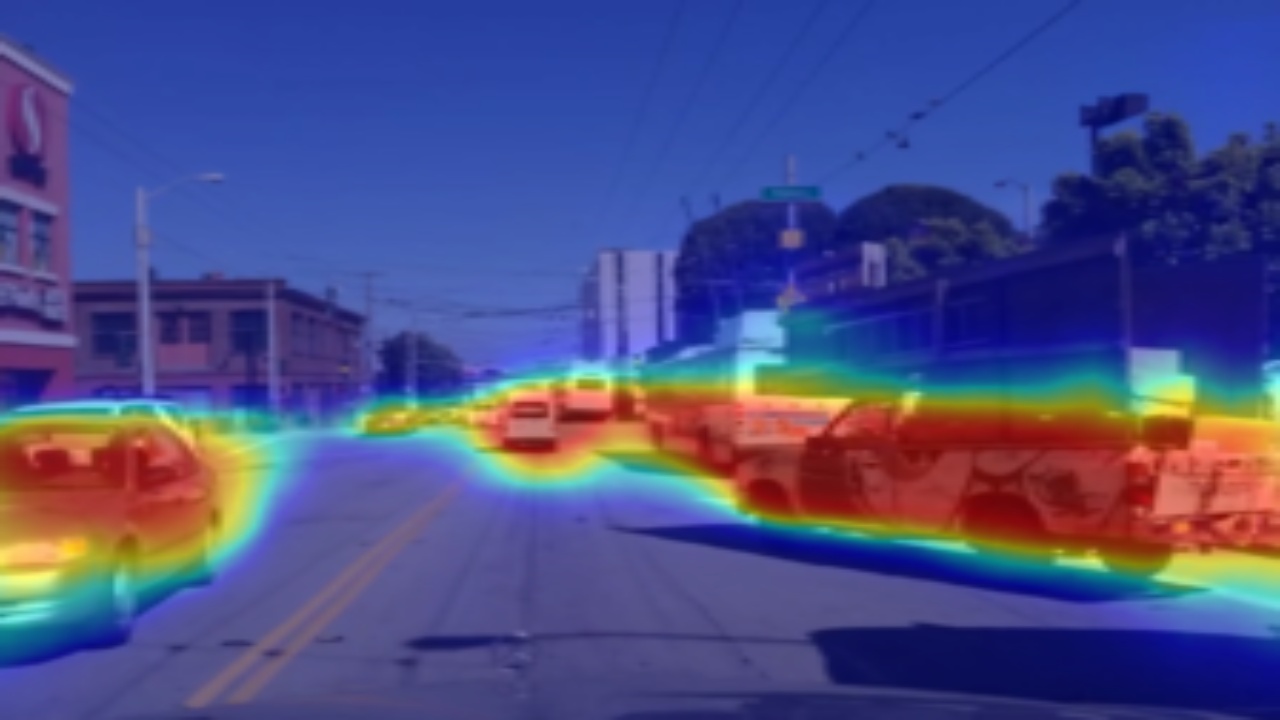} }}%
    \quad
    \caption{Comparison of the prediction of four popular saliency models trained on the BDD-A ground-truth (middle row) and our SAGE groundtruth (bottom row). It can be seen that for each model, SAGE trained results can capture more detailed semantic context (Best viewed in color).}%
    \label{fig:model-comparison}%
\end{figure*}

%% file: files/algorithm.tex
\makeatletter
\def\BState{\State\hskip-\ALG@thistlm}
\makeatother

\begin{algorithm}
\caption{SAGE-Net($\mathbf{\mathcal{X}_{RGB}}, v_{thresh}$)}\label{euclid}
\begin{algorithmic}[1]
\State $\mathcal{Y}_{\text{RGB}}$ $\gets \mathrm{Saliency}$ $\mathrm{model}(\mathbf{\mathcal{X}_{RGB}})$
\State $\mathcal{X}_{\text{last}} \gets \mathbf{\mathcal{X}_{RGB}}[-1]$
\State $\mathcal{D}_{\text{RGB}}$ $\gets \mathrm{Monodepth2}(\mathcal{X}_{\text{last}})$
\State $\mathcal{Y}_{\text{RGB-D}}$ $ \gets \mathcal{Y}_{\text{RGB}}$ $\oplus$ $\mathcal{D}_{\text{RGB}}$
\If {$v_{ego}(\mathcal{X}_{\text{last}}) > v_{thresh}$} \Return $\mathcal{Y}_{\text{RGB-D}}$
\Else
\State $\mathcal{I}_{\mathcal{X}_{\text{last}}} \gets \mathrm{ResEnDec}(\mathbf{\mathcal{X}_{RGB}})$
\If {$\mathcal{I}_{\mathcal{X}_{\text{last}}} = \mathtt{not}$ $\mathtt{crossing}$} \Return $\mathcal{Y}_{\text{RGB-D}}$
\Else
\State $\mathbf{bbox} \gets \mathcal{O}(\mathcal{X}_{\text{last}})$
\State $\mathcal{Y}_{\text{final}} \gets \mathcal{Y}_{\text{RGB-D}}$ $\otimes$ $\mathbf{bbox}$
\State \Return $\mathcal{Y}_{\text{final}}$
\EndIf
\EndIf
\end{algorithmic}
\label{algorithm}
\end{algorithm}

%% file: files/expt_result.tex
\section{Experiments and Results} \label{sect:expt_res}
Due to the simplicity of computation of our proposed ground-truth, several experiments can be run using it. These experiments can be split into a two-stage hierarchy - (i) conducted over the entire dataset comprising of multiple combinations in driving scenarios - day vs night, city vs countryside, intersection vs highway etc. and (ii) those over specific important driving conditions such as intersection regions and crowded streets. The reason for the latter set of experiments is that averaging out the predicted results over all scenarios is not always reflective of the most important situations requiring maximum human attention \cite{xia2018predicting}. For all the experiments, we describe the evaluation metrics used for comparison, and using those, compare the results of the gaze-only groundtruth and our proposed SAGE groundtruth for the different algorithms and datasets.
\subsection{Some popular saliency prediction algorithms} \label{algos}
We selected four popular saliency prediction algorithms from an exhaustive list for training with SAGE groundtruth and compared their performance against those trained with gaze-only maps. The first set of algorithms, DR(eye)VE \cite{palazzi2017learning} and BDD-A \cite{xia2018predicting}, were created exclusively for saliency prediction in the driving context. For DR(eye)VE, we only consider the image-branch for our analysis instead of the multi-branch network \cite{palazzi2018predicting} due to two main reasons which make real-time operation possible. Firstly, it has a fraction of the number of trainable parameters and hence is faster to train and evaluate. Secondly, the latter assumes that the optical flow and semantic segmented maps are pre-computed even at test time, which is difficult to achieve online. The BDD-A algorithm is more compact and it consists of a visual feature extraction module \cite{krizhevsky2012imagenet}, followed by a feature and temporal processing unit in the form of 2D convolutions and Convolutional LSTM (Conv2D-LSTM) \cite{xingjian2015convolutional} network respectively. However, both these algorithms combine the features extracted from the final convolution layers to make the saliency maps. This mechanism ignores low-level intermediate representations such as edges and object boundaries, which are important detections for driving scenario. Thus, we also consider ML-Net \cite{mlnet2016}, which achieved best results on the largest publicly available image saliency datset SALICON \cite{jiang2015salicon}. It extracts low, medium, and high-level image features and generates a fine-grained saliency map from them. Finally, PiCANet \cite{liu2018picanet} extends this notion further by generating an attention map at each pixel over a context region and constructing an attended contextual feature to further enhance the feature representability of ConvNets. Figure \ref{fig:model-comparison} shows a comparison of the predicted saliency maps trained on gaze-only ground-truth, and those obtained from SAGE. For nearly every gaze-only model, the focus of attention is entirely towards the center of the image, thereby ignoring other cars. In contrast, SAGE-trained models have managed to successfully capture this vital information. We refer the reader to Appendix B of the supplementary material for implementation details of these four algorithms.
\subsection{Evaluation metrics}

We consider a set of metrics which are suitable for evaluating saliency prediction in the driving context, as opposed to general saliency prediction. More specifically, for driving purpose, we want to be more careful about identifying "False Negatives (FN)" than "False Positives (FP)", since the former error holds a much higher cost. As illustrated in Section \ref{sect:methodology}, our proposed ground-truth has both a gaze component and a semantic component. Thus, we classify the set of metrics broadly into two categories - (i) fixation-centric and (ii) semantic-centric.

For the first category, we choose two distribution-based metrics - Kullback-Leibler Divergence ($\mathrm{D_{KL}}$), and Pearson's Cross Correlation ($\mathrm{CC}$). $\mathrm{D_{KL}}$ is an asymmetric dissimilarity metric, that penalizes FN more than FP. $\mathrm{CC}$, on the other hand is a symmetric similarity metric which equally affects both FN and FP, thus giving an overall information regarding the misclassifications that occurred. Another variant of fixation metrics are the location-based metrics, such as Area Under ROC Curve ($\mathrm{AUC}$), Normalized Scanpath Saliency ($\mathrm{NSS}$) and Information Gain ($\mathrm{IG}$), which operate on the ground-truth being represented as discrete fixation locations \cite{bylinskii2018different}. But for the driving task, it is crucial to identify every point on a relevant object, especially their boundaries, in order to mitigate risks. Thus, continuous distribution metrics are more appropriate here as they can better capture object boundaries.
\input{tables/model_comparison.tex}
\input{images/bar_plots.tex}

In the second category, we again consider two metrics - namely $\mathrm{F}$-$\mathrm{score}$, which measures region similarity of detection, and Mean Absolute Error ($\mathrm{MAE}$), which gives pixel-wise accuracy. $\mathrm{F}$-$\mathrm{score}$ is given by the formulae,
\begin{equation}
    F_{\beta} = \frac{(1 + \beta^2) * precision*recall}{\beta^2 *  precision + recall}
\end{equation}
where $\beta^2$ is a parameter that weighs the relative importance of precision and recall. In most literatures \cite{wang2019learning, achanta2009frequency, li2014secrets}, $\beta^2$ is taken to be 0.3, thus giving a higher weightage to precision. However, following the earlier discussion regarding varying costs associated with FN and FP for the driving purpose, we consider $\beta^2$ to be $1$, thereby assigning equal weightage to each. For a formal proof of this, we refer the reader to Appendix A of the supplementary material.
\subsection{Results and Discussion}

In this section, we discuss the experiments and results of algorithms trained on our proposed SAGE ground-truth, along with how they compare to the performance of the same algorithms, when trained on existing gaze-only ground-truths \cite{Alletto_2016_CVPR_Workshops, xia2018predicting}. We compare our results with that of BDD-A gaze in most of the experiments, since it is more reflective of scene semantics than the DR(eye)VE gaze. For fair comparison, we adopt different strategies for evaluating the fixation centric and semantic centric metrics. Since both the traditional gaze-only approach and SAGE contain gaze information, we use the respective ground-truths to evaluate the fixation metrics (\ie gaze for the gaze-only trained model, and SAGE for our trained model). However, for the semantic metrics, we use the segmented maps generated by Mask RCNN as ground-truth to evaluate how well each of the methods can capture semantic context. The first set of comparisons, given by Table \ref{tab: model comparison} and Figure \ref{fig:barplots}, are calculated by taking the average over the entire test set, while the remaining comparisons are for a subset of the test set involving two important driving scenarios, namely - pedestrians crossing at an intersection in Table \ref{tab: pred crossing}, and cars approaching towards the ego-vehicle in Table \ref{tab: car approaching}.

\textbf{Overall comparison} - In Table \ref{tab: model comparison}, we train the four algorithms described in \S\ref{algos} on the BDD-A dataset \cite{xia2018predicting}. We show the results obtained when evaluating the algorithms trained on the gaze-only data, and then on SAGE data generated by combining semantics with the gaze of \cite{xia2018predicting}. As observed from the table, the $\mathrm{D_{KL}}$ and $\mathrm{F_1}$ values obtained on SAGE are optimal for almost all the algorithms, while for $\mathrm{CC}$ and $\mathrm{MAE}$, it either performs better or is marginally poorer in performance. Overall, this analysis shows that our proposed SAGE ground-truth performs good on a diverse set of algorithms, thus proving its flexibility and robustness.

We next consider Figure \ref{fig:barplots}, where a cross-evaluation of our method with respect to different driving datasets is performed. For this set of experiments, we fix one algorithm, namely DR(eye)VE \cite{palazzi2017learning}, while we vary the data. We evaluate two variants of SAGE - first, by combining scene semantics with the gaze of \cite{Alletto_2016_CVPR_Workshops}, and second, with the gaze of \cite{xia2018predicting}. For each of these, we compare with the respective gaze-only ground-truth of the respective datasets. Like before we evaluate the performance of predicted saliency maps using the same fixation-centric and semantic-centric metrics. The results show that the proposed SAGE models are not strongly tied to a dataset and can adapt to different driving conditions. It is important to note that even though the cross evaluation (SAGE-D tested on \cite{xia2018predicting}, and SAGE-B tested on \cite{Alletto_2016_CVPR_Workshops}) is slightly unfair, the results for SAGE still significantly outperforms those of the respective gaze-only models.
\input{tables/newmetrics_ped.tex}
\input{tables/newmetrics_ca.tex}

\textbf{Comparison at important driving scenarios} - In Table \ref{tab: pred crossing}, we consider the scenarios of pedestrians crossing at intersections. For this purpose, we used a subset of the JAAD dataset \cite{kotseruba2016joint} containing more than five pedestrians (not necessarily as a group) crossing the road. The same four algorithms described in \S\ref{algos} have been reconsidered for this case. Using M-RCNN, the segmented masks of all the crossing pedestrians were computed and the predicted saliency maps from the models were evaluated against this baseline. Upon comparison, it can be seen that models trained on SAGE surpass those trained on the gaze-only ground-truth. It is to be noted that even though none of the models were trained on the JAAD dataset \cite{kotseruba2016joint}, the results are still pretty consistent across all the algorithms. This shows that learning from SAGE indeed yields a better saliency prediction model which can detect pedestrians crossing at an intersection more reliably.

Finally, in Table \ref{tab: car approaching}, we took into account another important driving scenario where we consider the detection of number of cars approaching the ego-vehicle as a metric. The evaluation set was constructed by us from different snippets of the DR(eye)VE \cite{Alletto_2016_CVPR_Workshops} and the BDD-A \cite{xia2018predicting} datasets, where a single or a group of cars is/are approaching the ego-vehicle from the opposite direction in an adjacent lane. Once again, we evaluated the four algorithms on this evaluation set. Like in Table \ref{tab: pred crossing}, here too, we analyze the detections with respect to those made by M-RCNN. The results from Table \ref{tab: car approaching} show that for almost each experiment the performance of algorithms trained on SAGE is consistent in detecting the vehicles more accurately compared to the models trained by gaze-only ground-truth.

To summarize, the experiments clearly show that the proposed SAGE ground-truth can be easily trained using different saliency algorithms and the obtained results can also operate well across a wide range of driving conditions. This makes it more reliable for the driving task as compared to existing approaches which only rely on raw human gaze. Overall, the performance of our method is better than gaze-only groundtruth on \textbf{49/56 (87.5\%)} cases, not only when averaged over the entire dataset, but more importantly, in specific driving situations demanding higher focus of attention.

%% file: tables/model_comparison.tex
\begin{table*}[!t]
\begin{center}
\small
\ra{1.2}
\scalebox{0.96}{
\begin{tabular}{lccrccrccrcc}\toprule
\multirow{2}{*}{} & \multicolumn{5}{c}{Fixation-centric metrics} && \multicolumn{5}{c}{Semantic-centric  metrics} \\
\cline{2-6} \cline{8-12}
 & \multicolumn{2}{c}{$\mathrm{D_{KL}}$} && \multicolumn{2}{c}{$\mathrm{CC}$} && \multicolumn{2}{c}{$\mathrm{F_1}$ $\mathrm{score}$} && \multicolumn{2}{c}{$\mathrm{MAE}$}\\ \cline{2-3} \cline{5-6} \cline{8-9} \cline{11-12}
Model & Gaze gt & \textbf{SAGE gt} && Gaze gt & \textbf{SAGE gt} && Gaze gt & \textbf{SAGE gt} && Gaze gt & \textbf{SAGE gt} \\
DREYEVE \cite{palazzi2018predicting} & 1.28$\pm$0.43 & \textbf{0.73$\pm$0.38} && 0.58$\pm$0.13 & \textbf{0.75$\pm$0.13} && 0.1$\pm$0.06 & \textbf{0.37$\pm$0.14} && 0.11$\pm$0.06 & \textbf{0.08$\pm$0.05}\\
BDDA \cite{xia2018predicting} & 1.34$\pm$0.67 & \textbf{1.02$\pm$0.49} && 0.54$\pm$0.23 & \textbf{0.6$\pm$0.18} && 0.12$\pm$0.11 & \textbf{0.46$\pm$0.19} && \textbf{0.12$\pm$0.09} & 0.13$\pm$0.07\\
ML-Net \cite{mlnet2016} & \textbf{1.1$\pm$0.32} & 1.35$\pm$0.51 && \textbf{0.64$\pm$0.13} & 0.6$\pm$0.14 && 0.12$\pm$0.07 & \textbf{0.43$\pm$0.14} && 0.12$\pm$0.06 & \textbf{0.1$\pm$0.06}\\
PiCANet \cite{liu2018picanet} & 1.11$\pm$0.28 & \textbf{0.83$\pm$0.31} && 0.64$\pm$0.11 & \textbf{0.73$\pm$0.11} && 0.15$\pm$0.08 & \textbf{0.64$\pm$0.15} && 0.11$\pm$0.06 & \textbf{0.11$\pm$0.05}\\
\bottomrule
\end{tabular}
}
\end{center}
\caption{Comparison of different saliency algorithms trained on BDD-A gaze gt and SAGE gt. All experiments are conducted on the BDD-A dataset.}
\label{tab: model comparison}
\end{table*}

%% file: images/bar_plots.tex
\begin{figure*}[!b] %
    \centering
    \subfloat[K-L Divergence ($\mathrm{D_{KL}}$)]{{\includegraphics[width=4cm]{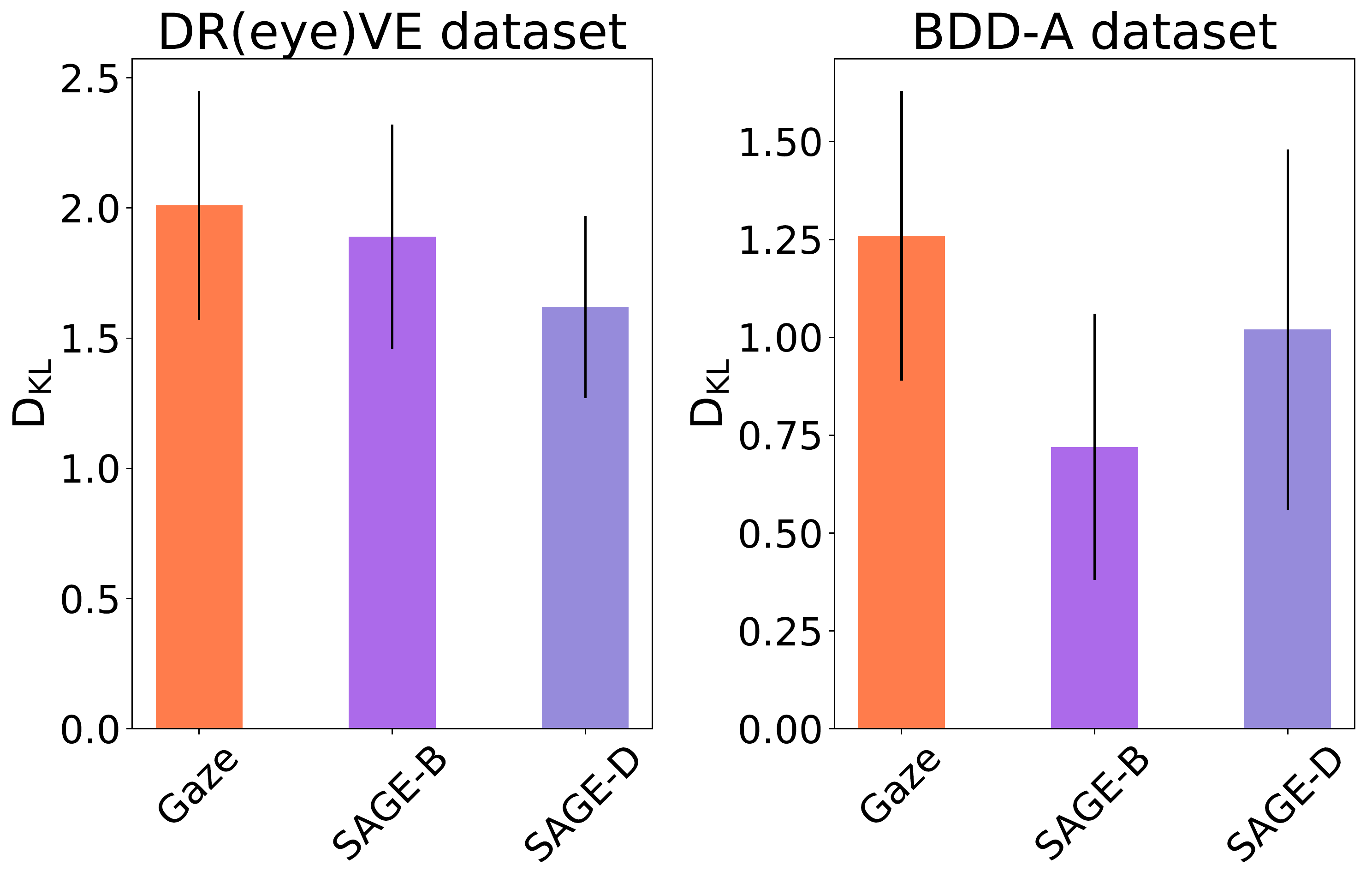} }}%
    \quad
    \subfloat[Cross Correlation ($\mathrm{CC}$)]{{\includegraphics[width=4cm]{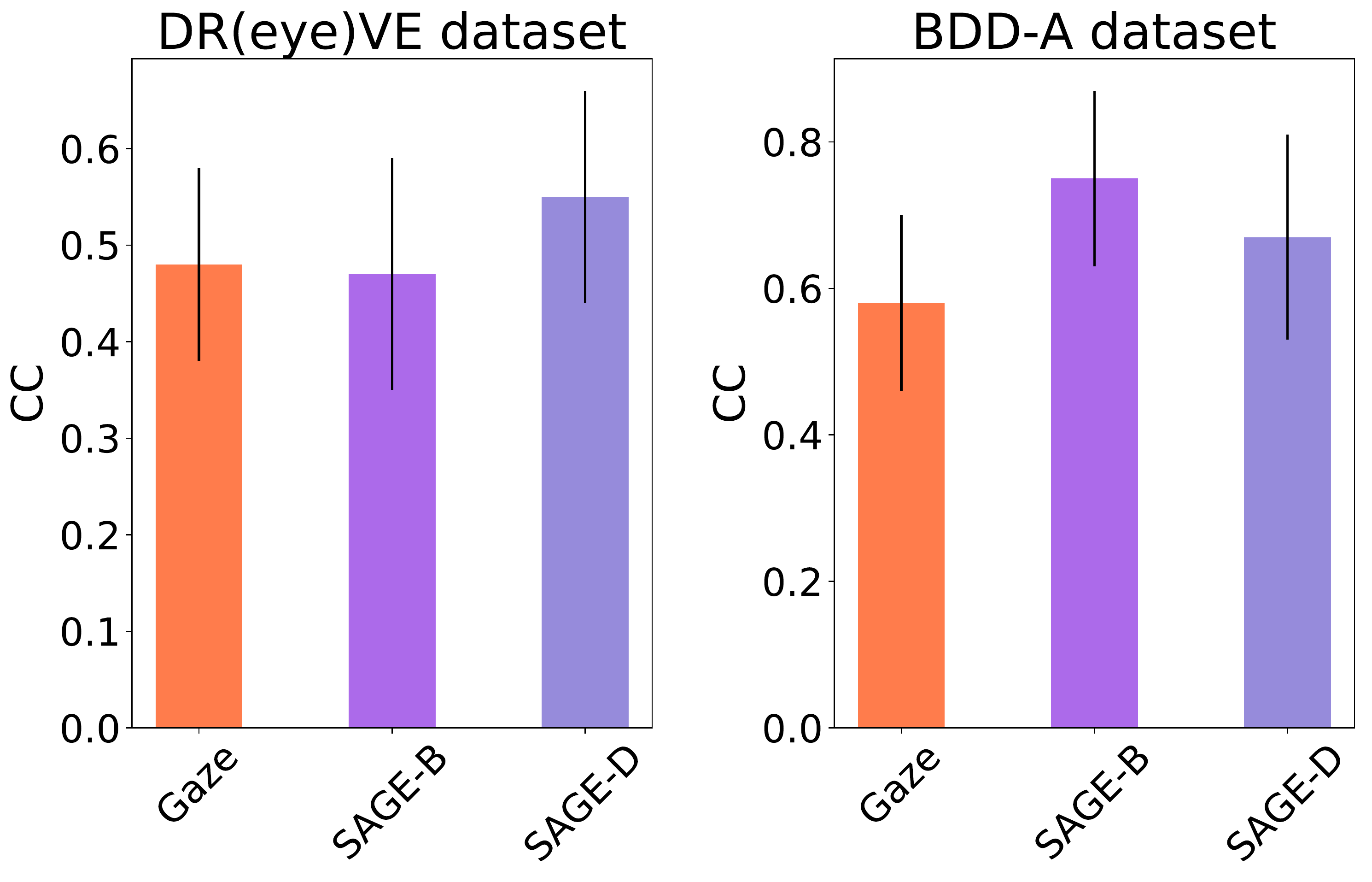} }}%
    \quad
    \subfloat[$\mathrm{F_1}$ $\mathrm{score}$]{{\includegraphics[width=4cm]{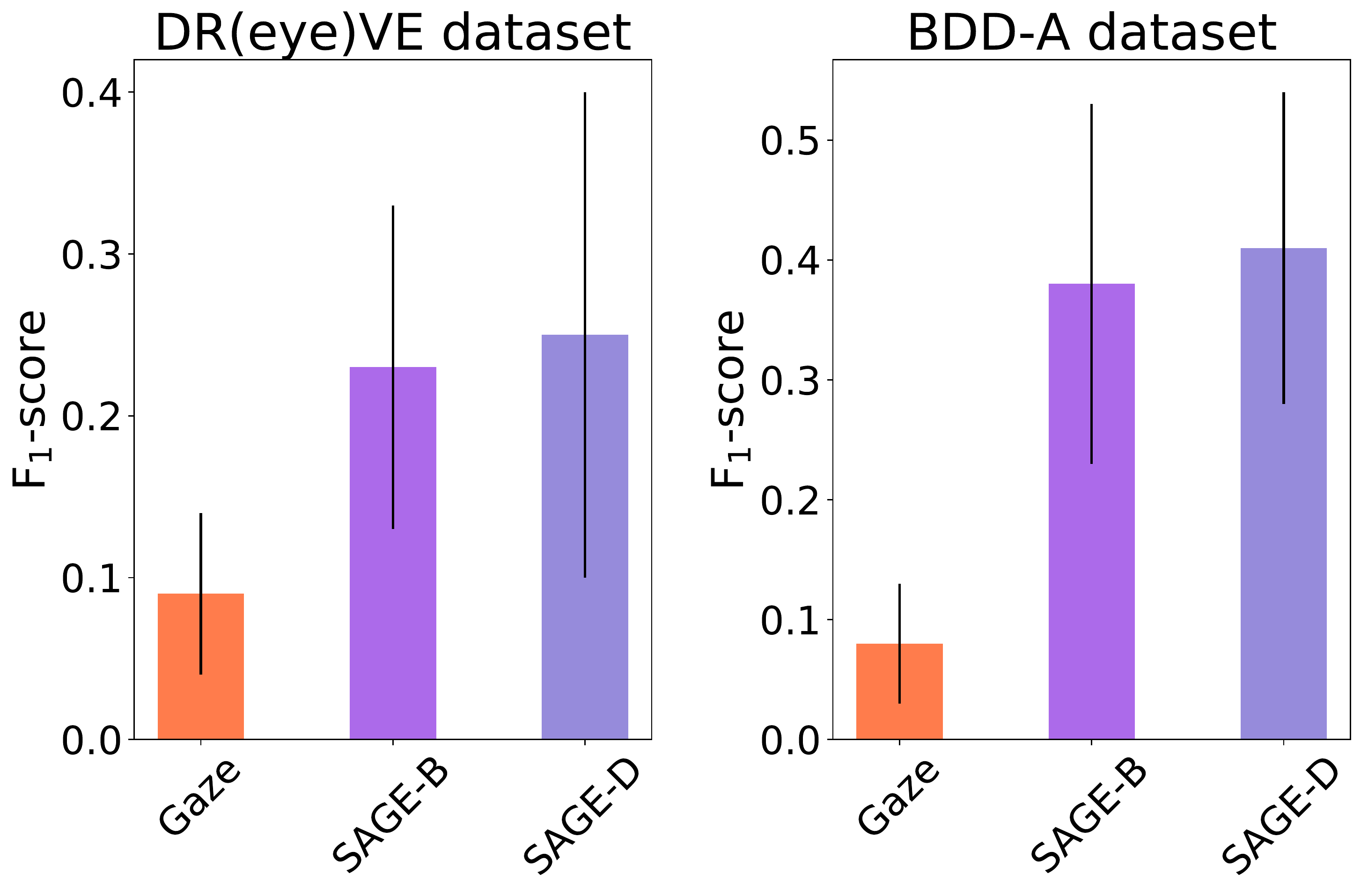} }}%
    \quad
    \subfloat[Mean Absolute Error ($\mathrm{MAE}$)]{{\includegraphics[width=4cm]{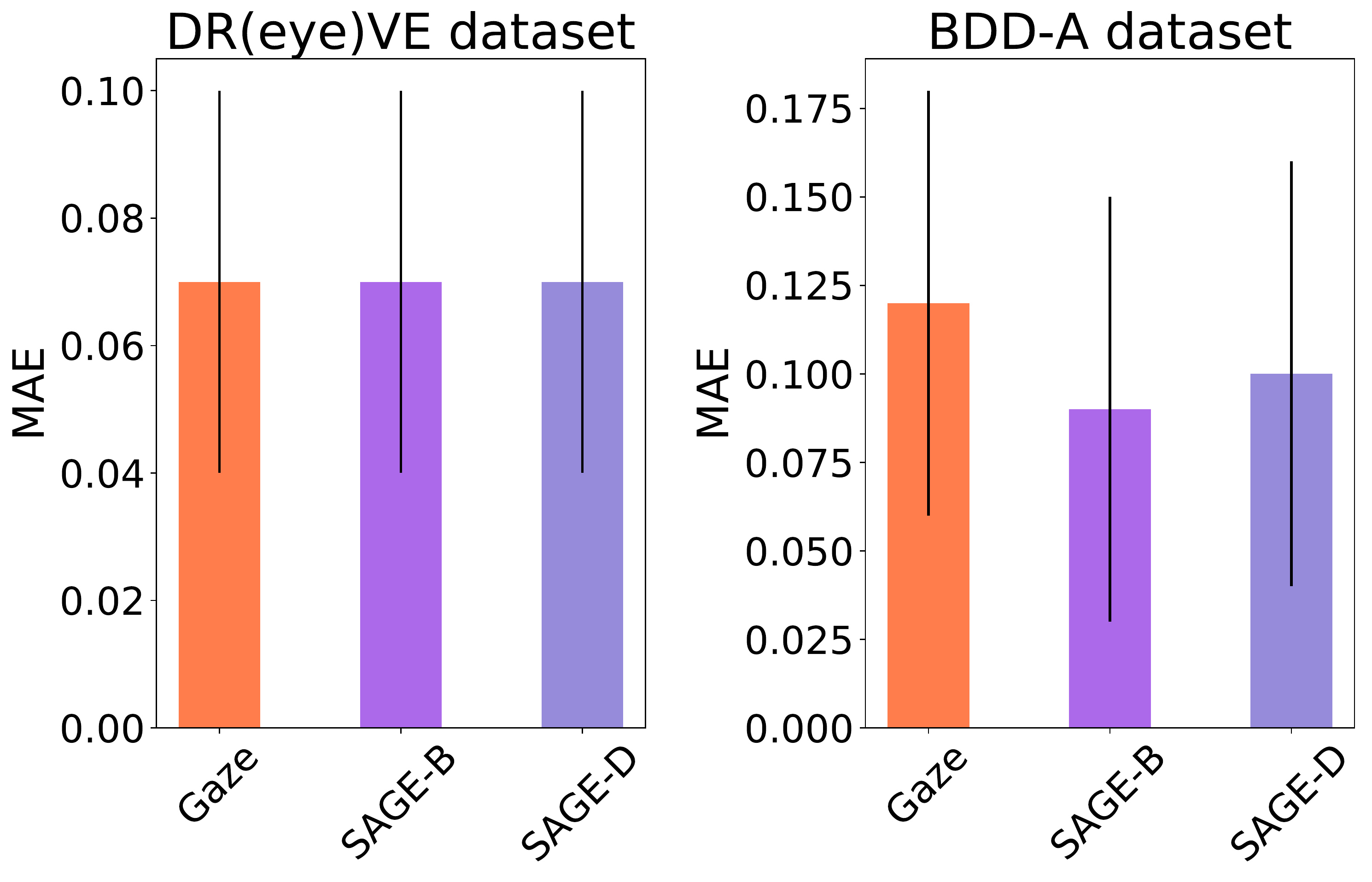} }}%
\caption{Cross-evaluation of SAGE-gt by considering the gaze of two different datasets. \cite{Alletto_2016_CVPR_Workshops} and BDD-A \cite{xia2018predicting} have been used for comparison. SAGE-B/D refers to the combination of semantics with the gaze of BDD-A/DR(eye)VE dataset.}%
    \label{fig:barplots}%
\end{figure*}

%% file: tables/newmetrics_ped.tex
\begin{table*}[!t]
\begin{center}
\small
\ra{1.1}
\scalebox{0.96}{
\begin{tabular}{lccrccrccrcc}\toprule
\multirow{2}{*}{} & \multicolumn{5}{c}{Fixation-centric metrics} && \multicolumn{5}{c}{Semantic-centric  metrics} \\
\cline{2-6} \cline{8-12}
 & \multicolumn{2}{c}{$\mathrm{D_{KL}}$} && \multicolumn{2}{c}{$\mathrm{CC}$} && \multicolumn{2}{c}{$\mathrm{F_1}$ $\mathrm{score}$} && \multicolumn{2}{c}{$\mathrm{MAE}$}\\ \cline{2-3} \cline{5-6} \cline{8-9} \cline{11-12}
Model & Gaze gt & \textbf{SAGE gt} && Gaze gt & \textbf{SAGE gt} && Gaze gt & \textbf{SAGE gt} && Gaze gt & \textbf{SAGE gt} \\
\hline
DREYEVE \cite{palazzi2018predicting} & 3.36$\pm$0.76 & \textbf{1.56$\pm$0.62} && 0.19$\pm$0.09 & \textbf{0.55$\pm$0.15} && 0.07$\pm$0.06 & \textbf{0.21$\pm$0.09} && 0.08$\pm$0.04 & \textbf{0.07$\pm$0.04}\\
BDDA \cite{xia2018predicting} & 2.37$\pm$0.78 & \textbf{1.87$\pm$0.81} && 0.28$\pm$0.16 & \textbf{0.43$\pm$0.16} && 0.2$\pm$0.13 & \textbf{0.37$\pm$0.17} && \textbf{0.09$\pm$0.05} & 0.12$\pm$0.04\\
ML-Net \cite{mlnet2016} & 2.44$\pm$0.58 & \textbf{2.27$\pm$0.67} && 0.29$\pm$0.11 & \textbf{0.41$\pm$0.15} && 0.15$\pm$0.07 & \textbf{0.31$\pm$0.13} && 0.09$\pm$0.04 & \textbf{0.08$\pm$0.04}\\
PiCANet \cite{liu2018picanet} & 2.97$\pm$0.68 & \textbf{1.81$\pm$0.72} && 0.20$\pm$0.11 & \textbf{0.50$\pm$0.14} && 0.13$\pm$0.07 & \textbf{0.44$\pm$0.16} && \textbf{0.07$\pm$0.04} & 0.11$\pm$0.03\\
\bottomrule
\end{tabular}
}
\end{center}
\caption{Comparison of SAGE with the gaze models for pedestrian crossing at intersection scenario. The clips are taken from the JAAD \cite{kotseruba2016joint} dataset.}
\label{tab: pred crossing}
\end{table*}

%% file: tables/newmetrics_ca.tex
\begin{table*}[!b]
\begin{center}
\small
\ra{1.1}
\scalebox{0.96}{
\begin{tabular}{lccrccrccrcc}\toprule
\multirow{2}{*}{} & \multicolumn{5}{c}{Fixation-centric metrics} && \multicolumn{5}{c}{Semantic-centric  metrics} \\
\cline{2-6} \cline{8-12}
 & \multicolumn{2}{c}{$\mathrm{D_{KL}}$} && \multicolumn{2}{c}{$\mathrm{CC}$} && \multicolumn{2}{c}{$\mathrm{F_1}$ $\mathrm{score}$} && \multicolumn{2}{c}{$\mathrm{MAE}$}\\ \cline{2-3} \cline{5-6} \cline{8-9} \cline{11-12}
Model & Gaze gt & \textbf{SAGE gt} && Gaze gt & \textbf{SAGE gt} && Gaze gt & \textbf{SAGE gt} && Gaze gt & \textbf{SAGE gt} \\
\hline
DREYEVE \cite{palazzi2018predicting} & 3.87$\pm$0.79 & \textbf{1.28$\pm$0.71} && 0.18$\pm$0.11 & \textbf{0.62$\pm$0.19} && 0.08$\pm$0.08 & \textbf{0.33$\pm$0.16} && 0.08$\pm$0.05 & \textbf{0.07$\pm$0.05}\\
BDDA \cite{xia2018predicting} & 2.95$\pm$0.96 & \textbf{1.92$\pm$1.01} && 0.19$\pm$0.16 & \textbf{0.42$\pm$0.18} && 0.14$\pm$0.13 & \textbf{0.34$\pm$0.19} && \textbf{0.09$\pm$0.09} & 0.12$\pm$0.07\\
ML-Net \cite{mlnet2016} & 2.72$\pm$0.6 & \textbf{1.94$\pm$0.9} && 0.21$\pm$0.1 & \textbf{0.5$\pm$0.18} && 0.12$\pm$0.07 & \textbf{0.37$\pm$0.14} && 0.09$\pm$0.05 & \textbf{0.08$\pm$0.05}\\
PiCANet \cite{liu2018picanet} & 3.17$\pm$0.6 & \textbf{1.69$\pm$0.88} && 0.18$\pm$0.1 & \textbf{0.55$\pm$0.17} && 0.12$\pm$0.07 & \textbf{0.49$\pm$0.2} && \textbf{0.08$\pm$0.05} & 0.1$\pm$0.04\\
\bottomrule
\end{tabular}
}
\end{center}
\caption{Comparison of SAGE with the gaze models for detecting multiple cars approaching the ego-vehicle from the opposite direction. The clips are taken from the DR(eye)VE \cite{Alletto_2016_CVPR_Workshops} and BDD-A \cite{xia2018predicting} datasets.}
\label{tab: car approaching}
\end{table*}

%% file: files/conclusion.tex
\section{Conclusion and Future Work} \label{sect:conclusion}

In this paper we introduced SAGE-Net, a novel deep learning framework for successfully predicting "where the autonomous vehicle  should look" while driving, through predicted saliency maps that learn to capture semantic context in the environment, while retaining the raw gaze information. With the proposed SAGE-groundtruth, saliency models have been shown to have attention on the important driving-relevant objects while discarding irrelevant or less important cues, without having any additional computational overhead to the training process. Extensive set of experiments demonstrate that our proposed method improves the performance of existing saliency algorithms across multiple datasets and various important driving scenarios, thus establishing the flexibility, robustness and adaptability of SAGE-Net. We hope that the research conducted in this paper will motivate the autonomous driving community into looking at strategies, that are simple but effective, for enhancing the performance of currently existing algorithms.

Our future work will involve incorporating depth in the SAGE-groundtruth and then having the entire framework to be trained end-to-end. Currently this could not be achieved due to low variance in the depth data, leading to overfitting. Another possible direction that is being considered is to explicitly add motion dynamics of segmented semantic objects in the surroundings in the form of SegFlow \cite{cheng2017segflow}. Work in this area is under progress as we are building a campus-wide dataset with these kind of annotations through visual sensors and camera-LiDAR fusion techniques.

%% file: files/acknowledgement.tex
\section*{Acknowledgements} \label{sec:acknowledgements}

The authors would like to thank Army Research Laboratory (ARL) W911NF-10-2-0016 Distributed and Collaborative Intelligent Systems and Technology (DCIST) Collaborative Technology Alliance for supporting this research.